\documentclass[twoside,11pt]{article}

\usepackage{amsmath,amssymb,amsthm,mathtools}
\let\savedproof\proof
\let\endsavedproof\endproof
\usepackage[preprint]{jmlr2e}
\let\proof\savedproof
\let\endproof\endsavedproof
\hypersetup{hidelinks,pdftitle={Overcoming a Learning Bottleneck in Impartial Games: A Circuit Complexity Approach},pdfauthor={Soren Riis},pdfkeywords={impartial games, NIM, circuit complexity, quantised inference, learning bottlenecks}}
\usepackage{tabularx}
\newcolumntype{Y}{>{\raggedright\arraybackslash}X}
\usepackage{graphicx}

\usepackage{microtype}
\usepackage{enumitem}
\usepackage{array}
\usepackage{tikz}
\usetikzlibrary{arrows.meta,positioning,decorations.pathmorphing,calc}
\graphicspath{{./}{./figures/}}
\bibpunct{(}{)}{;}{a}{,}{,}

\newcommand{\AC}{\mathrm{AC}}
\newcommand{\TC}{\mathrm{TC}}

\newcommand{\poly}{\mathrm{poly}}
\newcommand{\Heav}{\mathbf{H}}

\newcommand{\NN}{\mathsf{NN}}
\newcommand{\RNN}{\mathsf{RNN}}
\newcommand{\LTST}{\mathsf{LTST}}

\newcommand{\bits}{\{0,1\}}

\newcommand{\N}{\mathbb{N}}

\newcommand{\Qset}{\mathcal{Q}}

\numberwithin{theorem}{section}
\numberwithin{equation}{section}

\title{Overcoming a Learning Bottleneck in Impartial Games:\\
A Circuit Complexity Approach}

\author{\name S{\o}ren Riis \email s.riis@qmul.ac.uk \\
\addr Centre for Fundamentals of AI and Computational Theory\\
Queen Mary University of London\\
Mile End Road, London E1 4NS, United Kingdom}
\ShortHeadings{Learning Bottlenecks in Impartial Games}{Riis}
\firstpageno{1}

\begin{document}
\maketitle

\begin{abstract}
Why can neural networks struggle to learn optimal play in a game with a simple, efficient solution algorithm? Motivated by empirical difficulties with parity in impartial games, we use NIM to study a representational bottleneck and three ways around it. We show that a restricted model of shallow, fixed-scale quantised networks lies in $\AC^0$, a circuit class unable to compute parity on arbitrarily large inputs. This prevents optimal single-position play across all board sizes within the model.
The obstruction also guides architectural design: two-frame inputs allow restoration from a known zero-nim-sum position; complementary rollout policies reveal the missing invariant; and persistent recurrence accumulates it row by row. Each route retains shallow network components while supplying additional information or processing time. Supervised experiments illustrate the benefits of two-frame restoration and multiple rollout heads; recurrence is a theoretical construction. These results establish representational possibilities, not guarantees of successful training: $\AC^0$ does not characterise practical learnability.
Core mathematical results and constructions have been checked in Lean, with classical complexity results as explicit external inputs. The interactive edition and complete Lean sources are available online \citep{riis2026interactive,riis2026lean}.
\end{abstract}



\section{Introduction}
\label{sec:intro}

Game play has historically served as the ``Formula 1'' of artificial intelligence: a high-speed, controlled proving ground where engineering limits are tested and new paradigms are born.
From Deep Blue in chess to AlphaGo in Go, games have repeatedly driven advances in search, planning, and reinforcement learning \citep{silver2016alphago,silver2017alphagozero,silver2018alphazero,schrittwieser2020muzero}.
Minsky and Papert's analysis of perceptron limitations illustrates how representational restrictions can motivate architectural changes \citep{minsky1969perceptrons}.

In NIM, two players take turns removing any positive number of tokens from a single heap; the player taking the last token wins. It belongs to the broad class of \emph{impartial games}: both players have the same legal moves from a given position. Many such board games can be thought of as games with shared pieces, whereas in \emph{partisan games}, such as chess, legal moves depend on which player is to move. NIM is central to impartial-game theory because every finite terminating impartial position under the last-move-wins convention is equivalent, in sums of games, to a single NIM heap (Section~\ref{sec:impartial}).

NIM poses a particularly revealing challenge: optimal moves can be found by a simple, efficient algorithm, yet learning that algorithm can be difficult. The puzzle is the gap between the ease of computing the solution once the rule is known and the difficulty of discovering it through training. Friedman's 2017 AI challenge asked whether a system could discover perfect NIM play without being given the winning strategy \citep{friedman2017challenge}.

Our earlier work with Bei Zhou provides the empirical starting point. In \emph{Impartial Games: A Challenge for Reinforcement Learning}, Zhou and Riis found that AlphaZero-style learning deteriorated as NIM boards grew \citep{zhou2022impartial}. Subsequent parity experiments examined how noisy labels can obstruct curriculum learning \citep{zhou2023parity}; see also Zhou's thesis \citep{zhou2024phd}. These findings make parity a useful test of the gap between learning familiar patterns and discovering an exact rule that works across positions.

Here we turn that difficulty into a question about computation. The class $\AC^0$ consists of polynomial-size Boolean circuits built from AND, OR and NOT gates, with a number of layers that remains fixed as the input grows. Such circuits can combine many local tests, but cannot compute global parity: whether the total number of input 1s is odd or even. This is the feature that makes the class useful here: it provides a precise model with a provable version of the parity limitation seen in the experiments. We establish the connection for an explicitly restricted neural model, rather than assuming that all neural networks lie in $\AC^0$.

The value of identifying a bottleneck is that it tells us what an architectural change must supply. Instead of asking a shallow network to rediscover a global invariant from every board, we can expose a useful transition, obtain complementary information from rollouts, or accumulate the invariant over time. Our contribution is this \emph{barrier-to-bypass} approach: a computational obstruction guides three concrete constructions. Representability in $\AC^0$ is a necessary condition for a fixed-latency computation in our restricted model; it is neither sufficient for successful training nor a universal condition on what neural networks can learn.

NIM itself does not depend on game history: the current position completely determines its legal moves and optimal play. History can nevertheless supply useful \emph{computational context} to a restricted network. The first and third bypasses use such context in different ways: an earlier board exposes a local change, while recurrent memory retains information accumulated during row processing and across positions.

The central observation is that while the \emph{state} can be hard to process, the \emph{transition} can be easy: under the promise that the preceding position has nim-sum zero, short history converts winning-move selection into local difference computation and restoration.
A second route uses distinct rollout policies to extract complementary bits of the invariant.
A third route processes each position sequentially, retaining recurrent state as rows and then successive positions arrive. Here a sequence of simple updates computes the global invariant, and memory connects the new position to the preceding one.
These mechanisms use different resources: explicit transition information, multiple rollout channels, or a growing recurrent computation horizon.
The distinction between a shallow component and the complete computation is essential. Two-frame restoration is a shallow map on a promised pair of boards. In the other two constructions, each policy step or recurrent update is shallow, but generating the information needed for a decision takes a sequence of such steps. Thus the bypasses preserve simple network components while changing the input information or total processing time. The complete sequential computation need not itself lie in $\AC^0$.
Figure~\ref{fig:barrier-blueprint} summarises the three routes for NIM.

\begin{figure}[t]
\centering
\begin{tikzpicture}[
    node distance=0.8cm and 1.2cm,
    auto,
    block/.style={
        rectangle,
        draw,
        fill=blue!5,
        text width=2.2cm,
        align=center,
        rounded corners,
        minimum height=1.0cm,
        font=\small
    },
    head/.style={
        rectangle,
        draw,
        fill=yellow!10,
        text width=1.8cm,
        align=center,
        rounded corners,
        minimum height=0.6cm,
        font=\footnotesize
    },
    line/.style={
        draw,
        -{Latex[length=2.5mm]},
        thick,
        gray!80
    },
    barrier/.style={
        circle,
        draw=red!80,
        fill=red!10,
        very thick,
        minimum size=0.6cm,
        inner sep=0pt
    },
    success/.style={
        circle,
        draw=green!60!black,
        fill=green!10,
        very thick,
        minimum size=0.6cm,
        inner sep=0pt
    },
    labelnode/.style={
        font=\bfseries\small,
        anchor=west
    }
]

    \node[block] (input1) {Input $P_t$\\ \textit{(Global)}};
    \node[block, right=of input1] (net1) {Single-Head\\ Network};
    \node[barrier, right=of net1] (out1) {\Large $\times$};
    \node[right=0.1cm of out1, align=left, font=\footnotesize, text width=2.5cm] {Cannot compute\\ global parity\\ in this model};

    \path[line] (input1) -- (net1);
    \path[line] (net1) -- (out1);
    \node[above=0.45cm of input1, labelnode] {(a) Barrier: Single-Frame, Single-Head};

    \node[block, below=1.3cm of input1] (input2) {Input $(P_{t-1}, P_t)$\\ \textit{(Transition)}};
    \node[block, right=of input2] (net2) {Single-Head\\ Network};
    \node[success, right=of net2] (out2) {\Large $\checkmark$};
    \node[right=0.1cm of out2, align=left, font=\footnotesize, text width=2.5cm] {Computes Local\\ Difference $\Delta$};

    \path[line] (input2) -- (net2);
    \path[line] (net2) -- (out2);
    \node[above=0.45cm of input2, labelnode] {(b) Bypass 1: Multi-Frame Input};

    \node[block, below=1.3cm of input2] (input3) {Input $P_t$\\ \textit{(Global)}};

    \node[right=of input3, text width=0cm, inner sep=0] (split) {};
    \node[head, right=0.2cm of split, yshift=0.4cm] (head1) {Head $\pi_0$};
    \node[head, right=0.2cm of split, yshift=-0.4cm] (head2) {Head $\pi_{B-1}$};
    \node[font=\footnotesize] at ($(head1)!0.5!(head2)$) {$\vdots$};

    \node[success, right=0.8cm of head1, yshift=-0.4cm] (out3) {\Large $\checkmark$};
    \node[right=0.1cm of out3, align=left, font=\footnotesize, text width=2.5cm] {Rollout outcomes\\ recover $B$ bits};

    \path[line] (input3) -- (split);
    \draw[line] (split.center) |- (head1.west);
    \draw[line] (split.center) |- (head2.west);
    \draw[line] (head1.east) -| (out3.north);
    \draw[line] (head2.east) -| (out3.south);

    \node[above=0.45cm of input3, labelnode] {(c) Bypass 2: Multiple Rollout Heads};

    \node[block, below=1.3cm of input3] (input4) {Rows of $P_t$\\ then $P_{t+1}$};
    \node[block, right=of input4] (net4) {Recurrent State\\ $s\leftarrow s\oplus x_i$};
    \node[success, right=of net4] (out4) {\Large $\checkmark$};
    \node[right=0.1cm of out4, align=left, font=\footnotesize, text width=2.5cm] {Nim-sum after\\ $N$ row updates\\ (growing latency)};
    \path[line] (input4) -- (net4);
    \path[line] (net4) -- (out4);
    \node[above=0.45cm of input4, labelnode] {(d) Bypass 3: Persistent Recurrence};
\end{tikzpicture}
\caption{The parity barrier and three structural bypasses.
(a) Fixed-scale quantised inference (FSQI) with fixed latency cannot compute global parity across arbitrarily large boards. Here fixed latency means a bounded number of sequential computation steps, independent of board size.
(b) Two frames expose $\Delta$ for restoration when the predecessor has nim-sum zero.
(c) Distinct policy heads recover complementary information through complete rollouts.
(d) Persistent recurrence accumulates the nim-sum row by row and retains summaries across positions. Each update has constant depth; the full scan has depth growing with $N$.}
\label{fig:barrier-blueprint}
\end{figure}

\subsection{Contributions}
\label{sec:contrib}

The results follow a single question: what must be added to a shallow learner to make the required strategy representable? We first identify the obstruction, then show how each bypass supplies a different missing resource.

\begin{itemize}[leftmargin=*]
\item \textbf{$\AC^0$ containment for fixed-scale quantised inference.}
We formalise a fixed-grid Boolean literal-threshold model and its bounded-unrolling recurrent extension. Both are simulable by $\AC^0$ circuits (Theorem~\ref{thm:sim-ac0}). An attention-style extension is covered only if its scoring, selection and aggregation operations can each be implemented by the permitted Boolean threshold gates. This is a restriction on the computation, not a requirement that every game have the same board size. Within a play sequence we use a fixed layout of $N$ heap rows and $B$ bits per heap, retaining empty heaps as zero rows. The asymptotic statements concern families of networks for growing input lengths $NB$.

\item \textbf{Single-frame impossibility for strong mastery.}
Under this model, no single-frame policy/value network can strongly master NIM in the worst case (Theorem~\ref{thm:singleframe-impossible}).

\item \textbf{Two-frame bypass via local differences and restoration.}
With two consecutive frames, the agent can compute $\Delta(P,P')$ in $\AC^0$ (Lemma~\ref{lem:nimber-diff-ac0}). When $P$ has nim-sum zero, an $\AC^0$ move selector restores zero after the opponent's move (Proposition~\ref{prop:restore}). This is conditional restoration mastery, not an arbitrary-start verifier.

\item \textbf{Planning analysis and the single-policy bottleneck.}
We analyse a restricted measurement interface inspired by rollouts, rather than a general model of MCTS.
With $B$ structured policies one can recover all nim-sum bits (Proposition~\ref{prop:rollout-sufficiency}). Root and one-heap-modification queries to one fixed member of that family expose only one unknown linear functional (Proposition~\ref{prop:single-policy-bound}). With an explicitly bounded readout, this yields the distributional bound in Theorem~\ref{thm:search-gap}.

\item \textbf{Persistent recurrence via sequential row updates.}
A constant-depth recurrent update accumulates the nim-sum over $N$ heap rows and retains a summary across position separators (Proposition~\ref{prop:streaming-nimsum}). With a replay of the current board or stored heap rows, this supports optimal move selection. The construction relaxes fixed total latency, not fixed-scale quantisation (Section~\ref{sec:recurrent-bypass}).

\item \textbf{Controlled empirical comparison and learnability discussion.}
On 20-heap, 4-bit NIM with $10^6$ supervised restoration examples, a one-frame model has low restoration accuracy while a two-frame model learns restoration (Section~\ref{sec:supervised-restoration}).
In a complementary single-frame experiment, multi-head FSM-controlled shootouts achieve perfect win/loss position classification while the single-head baseline remains bottlenecked by one-bit information (Section~\ref{sec:exp-fsm-shootout}).
We also discuss why $\AC^0$ is a useful abstraction for what is robustly feasible under fixed-scale quantised inference (Section~\ref{sec:majority-vs-parity}).
\end{itemize}

\subsection{Scope and limitations}
\label{sec:scope}

Our theorems are \emph{worst-case representational} statements about what can be computed under explicit architectural constraints.
They are not training-complexity results, and they intentionally abstract away from several practical details, such as continuous-valued activations, floating-point arithmetic, normalisation layers and the dynamics of optimisation.

\begin{itemize}[leftmargin=*]
\item \textbf{Fixed-scale quantisation / fixed latency.}
The $\AC^0$ simulation (Theorem~\ref{thm:sim-ac0}) assumes constant depth and a \emph{fixed-scale, bounded-range quantisation} regime: weights/thresholds lie on a fixed grid whose range/resolution do not scale with the input length (cf.\ Remark~\ref{rem:ac0-scope}).
Architectures that rely on fan-in dependent re-normalisation, biases/thresholds growing with dimension, or depth growing with $n$ fall outside this abstraction.
In particular, a fixed recurrent update rule need not have fixed total latency: the third bypass deliberately allows $N$ successive row updates per position.

\item \textbf{Formal model vs.\ experimental networks.}
The formal units have nonnegative weights on Boolean literals and bounded thresholds. The experiments instead use ordinary floating-point signed linear layers, ReLU and LayerNorm, with Log-Softmax in the loss. They test the benefit of transition information under the reported training protocol, not the asymptotic lower bound or the deployment behaviour of quantised networks.

\item \textbf{Stylised planning interface.}
The single-channel lower bound permits only root and one-heap-modification queries to a fixed structured policy, and requires a polynomial-size, constant-depth final readout. It is not a bound on arbitrary rollout policies, unrestricted query transformations, or an unbounded adaptive MCTS controller.

\item \textbf{No guarantee of learnability.}
The multi-policy-head and recurrent constructions concern \emph{existence} of suitable computations; we do not claim that self-play efficiently discovers the structured heads or an exact recurrent accumulator.
Our experiments use controlled supervised restoration and policy imitation; self-play discovery is left for a separate study.
\end{itemize}

\subsection{Paper organisation}
Section~\ref{sec:framework} introduces the fixed-scale quantised inference model and the $\AC^0$ simulation theorem.
Section~\ref{sec:related} situates our contribution in circuit complexity, modern sequence models, and reinforcement learning for combinatorial games.
Section~\ref{sec:impartial} recalls NIM and the nim-sum characterisation.
Section~\ref{sec:main} presents the main impossibility and bypass results: two-frame restoration, rollout channels (Section~\ref{sec:search-gap}), and persistent recurrence (Section~\ref{sec:recurrent-bypass}).
Section~\ref{sec:learnability} discusses modelling choices and learnability considerations.
Section~\ref{sec:experiments} reports experiments.
Appendices contain formal model definitions and proofs.

\section{Theoretical framework and neural network models}
\label{sec:framework}

\subsection{Fixed-scale quantised inference and \texorpdfstring{$\AC^0$}{AC0} simulation}
\label{sec:framework-fixedprecision}

We consider neural inference models that are \emph{(i)} bounded depth, \emph{(ii)} polynomial size, and \emph{(iii)} operate in a \emph{fixed-scale quantised} regime:
all weights and thresholds lie on a fixed quantisation grid of bounded dynamic range that does \emph{not} change with the input length.
The fixed scale is motivated by limits on the \emph{effective precision} of neural computation, including sensitivity to noise: distinctions that can be used reliably may be much coarser than the rounding resolution of 32-bit floating-point arithmetic. Noise motivates this abstraction; it does not by itself imply the gate and threshold assumptions below.

Our Boolean threshold gates sum nonnegative weights on inputs that may be negated. Negation allows inhibitory as well as excitatory effects: for a Boolean input $x$, the identity $-ax=a(1-x)-a$ replaces a negative coefficient by a positive weight on the negated input, with an adjustment to the threshold. The substantive restriction is that the resulting weights and thresholds remain on the fixed, bounded grid; this is not a model of arbitrary signed-weight or INT8/fixed-point inference.

Concretely, we fix constants:
\begin{itemize}[leftmargin=*]
\item a \emph{quantisation resolution} parameter captured by a fixed denominator $D\in\N$ (grid step $1/D$),
\item a \emph{dynamic-range} bound $W\in\N$ (maximum magnitude $W/D$),
\item and (for recurrent/attention models) a finite time window $T\in\N$.
\end{itemize}
We define
\[
\Qset_{W,D} \;=\; \Big\{ \frac{\ell}{D}\ :\ \ell\in\{0,1,2,\dots,W\}\Big\},
\]
a finite set independent of input length.
The threshold bound applies \emph{after} choosing the literal representation. Replacing a signed term $-a x$ by $a(1-x)$ raises the threshold by $a$. With many such terms, the adjusted threshold may grow with the input length. What matters is the amount of evidence needed to activate a gate: our model bounds the number of satisfied input literals that suffice, whereas an unrestricted majority gate needs a number that grows with its inputs. Negative coefficients themselves are not the obstruction; the bound on the adjusted threshold is essential to the simulation.

Each neuron computes a thresholded weighted sum of Boolean inputs. Formal definitions for $\NN$, $\RNN$ and $\LTST$ are in Appendix~\ref{app:defs}.

\begin{definition}[Quantised inference]
\label{def:quantised-inference}
A network $\NN$ in this class has constant depth and polynomially many threshold neurons, with weights and thresholds in $\Qset_{W,D}$.
Analogously, fixed-scale quantised $\RNN$ and $\LTST$ models have constant depth and polynomial size per time step and operate within a fixed time window $T$.
For the simulation theorem, $T$ bounds the entire unrolled computation from an input-independent initial state, including any computation used to form a retained state.
\end{definition}

A bound on the number of preceding states consulted by one update is insufficient: even $h_{r+1}=h_r\oplus x_r$ can accumulate parity over an arbitrarily long stream. Section~\ref{sec:recurrent-bypass} exploits precisely this distinction by allowing the number of recurrent updates to grow with the number of rows.

\paragraph{Why fixed-scale quantisation suggests an $\AC^0$ abstraction.}
The key structural property is \emph{fixed scale}: the smallest nonzero weight is $1/D$ and the largest threshold is $W/D$, with $W,D$ independent of the input length.
Consequently, whenever a threshold unit outputs $1$, there is a satisfying \emph{witness} consisting of at most $W$ input literals whose contributions already cross the threshold.
This yields a constant-width DNF expansion for each unit (Lemma~\ref{lem:unit-ac0}), and composing a constant number of layers yields an $\AC^0$ simulation (Theorem~\ref{thm:sim-ac0}).

We emphasise that this is not a statement about floating-point arithmetic per se.
The abstraction intentionally excludes \emph{scale-adaptive} regimes where thresholds/biases grow with fan-in or where weights shrink with dimension (e.g.\ $1/n$ or $1/\sqrt{n}$ normalisations), which move the analysis toward threshold-circuit classes such as $\TC^0$. This class allows constant-depth, polynomial-size circuits with general threshold gates, including majority gates; unlike $\AC^0$, it can compute parity.
\begin{remark}[Scope of the $\AC^0$ abstraction]
\label{rem:ac0-scope}
Theorem~\ref{thm:sim-ac0} formalises a fixed-latency \emph{fixed-scale quantised} regime in which each unit uses parameters drawn from a bounded-range grid $\Qset_{W,D}$ with $W,D$ independent of the input length.
This yields the constant-witness property exploited in Lemma~\ref{lem:unit-ac0}.

The abstraction is deliberately conservative: it does not model architectures that rely on fan-in dependent re-scaling, numerically large signed biases/thresholds that grow with dimension, or deep computations whose depth scales with $n$.
Allowing such scale-adaptive behaviour leads naturally to richer threshold-circuit models (e.g.\ $\TC^0$), in which parity-like computations can become representable.
\end{remark}

\begin{remark}[Local vs.\ global parity]
\label{rem:local-vs-global-parity}
When we say ``parity is hard'' we mean \emph{global} parity across the full Boolean input of length $n=NB$.
In contrast, parity on $O(1)$ bits is trivial in $\AC^0$ (hardwire the truth table), and more generally XOR of a constant number of $B$-bit blocks is $\AC^0$-computable.
This local/global separation underlies both the two-frame bypass via the nimber difference $\Delta(P,P')$ and the recurrent bypass, which composes local XOR updates over a growing number of steps.
\end{remark}

\begin{center}
\begin{tabular}{|>{\raggedright\arraybackslash}p{0.30\linewidth}|>{\raggedright\arraybackslash}p{0.28\linewidth}|>{\raggedright\arraybackslash}p{0.30\linewidth}|}
\hline
\textbf{Property of deployed inference} & \textbf{Captured here} & \textbf{Not captured here} \\
\hline
Fixed latency / bounded depth & constant depth $L$ (and window $T$) & depth growing with $n$ \\
\hline
Fixed-scale quantisation (bounded range) & weights/thresholds in fixed grid $\Qset_{W,D}$ & fan-in dependent scaling or thresholds growing with $n$ \\
\hline
Parallel bit operations & unbounded fan-in AND/OR/NOT & exact global counting / exact parity on unbounded inputs \\
\hline
Signed effects & via input negations (Lemma~\ref{lem:unit-ac0}) & large signed dot-product accumulation \\
\hline
\end{tabular}
\end{center}

\paragraph{Notation convention (input length).}
We write $N$ for the number of heaps and $B$ (or $k$ in experiments) for the bit-width used to encode each heap size. The resulting Boolean input length is $n := NB$. When we discuss $\AC^0$ circuit size/depth, $n$ always refers to this Boolean input length.

\paragraph{Circuit complexity primer.}
$\AC^0$ is the class of Boolean functions computable by constant-depth, polynomial-size circuits with unbounded fan-in AND/OR/NOT gates. A classical result is that $\mathrm{PARITY}\notin\AC^0$. In contrast, $\TC^0$ augments $\AC^0$ with unbounded fan-in threshold gates and can compute parity. Our focus here is a fixed-scale quantised regime in which thresholds do not scale with fan-in; in that regime, each threshold unit admits a constant-witness DNF expansion and the overall computation collapses to $\AC^0$ (Theorem~\ref{thm:sim-ac0}).

\subsection{Simulation by circuits}
\label{sec:simulation}

We state the simulation theorem for the literal-threshold model and its explicitly gate-realised extensions.

\begin{theorem}[Simulation by $\AC^0$]
\label{thm:sim-ac0}
Fix constants $W,D\in\N$ and a time window $T\in\N$.
Let $\{\mathcal{M}_n\}_{n\ge 1}$ be any family of models of type $\NN$, $\RNN$ (with window $T$), or $\LTST$ (with window $T$) such that:
\begin{enumerate}[leftmargin=*,label=(\arabic*)]
\item (\textbf{Polynomial size}) the number of neurons/units in $\mathcal{M}_n$ is at most $\poly(n)$.
\item (\textbf{Constant depth}) the number of layers is a fixed constant $L$ (independent of $n$); for $\RNN/\LTST$ the entire unrolled computation uses at most $T$ time steps from an input-independent initial state.
\item (\textbf{Permitted primitives}) every operation is realised by the Boolean literal-threshold gates of Definition~\ref{def:NN-formal}; all weights and thresholds lie in $\Qset_{W,D}$. For the attention-style extension this includes score computation, coefficient selection and aggregation.
\end{enumerate}
Then the Boolean function computed by $\mathcal{M}_n$ can be computed by an $\AC^0$ circuit family of polynomial size and constant depth (depending only on $L,T,W,D$).
\end{theorem}

\section{Related Work}
\label{sec:related}

\paragraph{Circuit complexity and neural networks.}
Parity is outside $\AC^0$ \citep{furst1984parity,hastad1986thesis}, while constant-depth threshold circuits ($\TC^0$) can compute it.
Attention results depend on the precise model: Hao, Angluin and Frank analyse hard-attention variants, including $\AC^0$ restrictions for unique hard attention \citep{hao2022hardattention}. Merrill and Sabharwal study the additional power of intermediate decoding over a $\TC^0$ baseline \citep{merrill2021transformer}; that containment does not rule out parity. Their sequential-computation perspective is relevant to our recurrent bypass, although the architectures and formal assumptions differ.
Our contribution applies these distinctions to NIM policies and a restricted rollout-measurement interface.

\paragraph{Quantised inference and deployment constraints.}
Low-bit inference motivates studying bounded numerical resources \citep{jacob2018integeronly,dettmers2022llmint8}. Our constant-witness proof additionally requires nonnegative literal weights, bounded thresholds and Boolean intermediate states. These are explicit mathematical assumptions, not properties inferred merely from a quantisation format. Standard signed linear layers and scale-adaptive normalisation are outside the theorem unless separately reduced to the permitted gates.

\paragraph{Reinforcement learning in combinatorial games.}
While RL has mastered domains such as Go and chess, impartial games remain a useful stress test because optimality is governed by exact algebraic invariants rather than approximate pattern matching.
Zhou and Riis \citep{zhou2022impartial,zhou2023parity} studied reinforcement-learning difficulties in impartial games.
Our lower bound identifies a representational obstruction in a restricted model; determining which mechanisms explain failures in ordinary neural training requires separate experiments.

\paragraph{History and memory in RL.}
Using short history windows is standard in partially observable settings, where stacked frames recover hidden variables such as velocity \citep{mnih2015humanlevel}.
Our point is different and perhaps surprising: even in Markovian, fully observable games---where the current state is informationally sufficient and the past is theoretically irrelevant---history can reduce \emph{computational} complexity by exposing local differences that would otherwise require expensive search or unbounded-depth inference to extract from the current state alone.
In our restoration setting, the previous frame carries a promise established by an earlier successful move: its nim-sum is zero. That promise makes a local difference sufficient for the next response. Establishing the promise initially remains a separate computational task.

\section{Impartial games and NIM}
\label{sec:impartial}

We briefly recall standard definitions; see \citep{berlekamp2001winningways} for extensive background.

\subsection{Impartial games}
\begin{definition}[Impartial game]
\label{def:impartial}
A finite terminating impartial game is a pair $(\mathcal{P},\mathcal{M})$ where $\mathcal{P}$ is a nonempty set of positions and $\mathcal{M}(p)\subseteq \mathcal{P}$ is the (finite) set of legal moves from $p$, with the property that the move relation does not depend on which player is to play (symmetry). We assume the move relation is acyclic and play terminates. We use the last-move-wins convention, usually called normal play: a player with no legal move loses.
\end{definition}

A position with no legal moves is terminal.

The last-move-wins convention also provides a convenient encoding of other decisive endings. A move that wins by capturing a designated piece or creating a specified configuration can simply lead to a position with no further legal moves. If instead reaching a special position loses for the player who made the move, give the opponent one forced move to a new terminal position. These encodings preserve the winner of the individual play; they must retain player-independent move rules to remain impartial. The theory below concerns the resulting normal-play game.

\subsection{NIM}
\begin{definition}[NIM]
\label{def:nim}
A NIM position consists of $h$ heaps with sizes $x_1,\dots,x_h\in\N_0$. A legal move reduces exactly one heap: choose $i$ with $x_i>0$ and replace $x_i$ by some $x_i'<x_i$.
\end{definition}

\subsection{Sprague--Grundy theory and nim-sum}
The breadth of impartial games makes the Sprague--Grundy theorem remarkable: every finite terminating impartial position under normal play is equivalent to a single NIM heap of some size, called its \emph{nimber} or Grundy value. Equivalence here means that replacing the position by that heap preserves the winning or losing status of its sum with any other such game. In a \emph{disjunctive sum}, a move is made in exactly one component. The nimber of a sum is the bitwise XOR of the component nimbers; in particular, the nimber of a NIM board is the XOR of its heap sizes.

\begin{definition}[Nim-sum]
\label{def:nimsum}
For heap sizes $x_1,\dots,x_h$, define
\[
\operatorname{nim}(x_1,\dots,x_h) \;=\; x_1 \oplus x_2 \oplus \cdots \oplus x_h,
\]
where $\oplus$ denotes bitwise XOR.
\end{definition}

\begin{theorem}[Classical NIM characterisation]
\label{thm:nim-win}
A NIM position $(x_1,\dots,x_h)$ is losing for the player to move iff $\operatorname{nim}(x_1,\dots,x_h)=0$. Moreover, from any winning position (nim-sum $\neq 0$) there is a move to a losing position (nim-sum $=0$).
\end{theorem}

\paragraph{Why NIM is a useful barrier example.}
For a disjunctive sum of independent impartial components, the Grundy value is the XOR of the component nimbers. If these nimbers can vary independently so that their aggregation contains an unbounded parity subproblem, NIM isolates the resulting aggregation obstruction.
Learning NIM would therefore address a fundamental component of impartial-game play, but would not by itself solve self-play learning for the wider class: determining component nimbers can be computationally difficult (Section~\ref{sec:othergames}). Nor does every impartial game contain a hard parity aggregation task. Our unconditional lower bound concerns NIM; extensions require an explicit encoding and reduction.

\section{Main results: impossibility and possibility}
\label{sec:main}

We first establish the single-frame barrier under fixed total latency, then develop three bypasses: recent-history inputs, multiple rollout channels, and persistent row-wise recurrence.

\subsection{Mastery notions}
We distinguish three notions of mastery:

\begin{definition}[Strong, weak, and restoration mastery]
\label{def:mastery}
An agent achieves \emph{strong mastery} of NIM if it chooses a winning move at every $N$-position where it is to move, including positions reached by arbitrary legal play by both players. Reachability is relative to the specified initial positions and player role; unrestricted strong mastery allows every initial position.

It achieves \emph{weak mastery relative to an initial position $p_0$ and a designated player role} if, whenever that role has a winning strategy from $p_0$, its own strategy wins against every opponent strategy. The guarantee concerns histories consistent with its own preceding actions, not every position reachable under arbitrary actions by both players. For asymptotic claims, the initial positions form a specified family. Strategies may depend on the complete history. Opponent moves are legal; a missing or illegal agent move counts as a loss. On an initially losing position the weak-mastery condition imposes no requirement.

It achieves \emph{restoration mastery} if, whenever the opponent moves from a $P$-position $Q$ with $\operatorname{nim}(Q)=0$ to a position $Q'$, the agent responds with a move $Q'\to R$ such that $\operatorname{nim}(R)=0$.
\end{definition}

We use standard impartial-game terminology: a $P$-position is losing for the player to move, and an $N$-position is winning.
For NIM, Theorem~\ref{thm:nim-win} says that $P$-positions are exactly the positions with nim-sum $0$. Equivalently, weak mastery requires the agent to move to a $P$-position at every one of its decision histories consistent with its own earlier actions, starting from an initial position winnable in its designated role. This distinction allows a simple strategy to master a family without solving positions it never encounters under its own play; Appendix~\ref{app:weak} gives examples.

\subsection{Single-frame impossibility results}
\label{sec:impossibility}

In a single-frame representation, the network sees only the current position. To play optimally, it must in effect determine whether the nim-sum is zero and, if not, find a move producing nim-sum zero. The core obstacle is that global nim-sum computation requires parity-like behaviour across many bits, and parity is not in $\AC^0$ \citep{furst1984parity,hastad1986thesis,smolensky1987algebraic,razborov1987lower}.

\begin{theorem}[Single-frame parity barrier]
\label{thm:singleframe-impossible}
Consider the restricted family of NIM positions $P(\mathbf{b})=(2,b_1,\dots,b_m)$ where $b_i\in\{0,1\}$.
Any single-frame agent that achieves strong mastery must compute the parity $b_1\oplus\cdots\oplus b_m$.
Since $\mathrm{PARITY}\notin \AC^0$, no polynomial-size, constant-depth fixed-scale quantised network can strongly master NIM.
\end{theorem}

\begin{proof}
Since each $b_i\in\{0,1\}$, these positions have nimber $2\oplus(b_1\oplus\cdots\oplus b_m)$, which is either $2$ or $3$. A winning move must change the first heap: changing any other heap leaves the leading binary digit of the nim-sum unchanged. To obtain nimber zero, the first heap must be reduced to $b_1\oplus\cdots\oplus b_m$. Thus the unique winning move removes two tokens when this parity is zero and one when it is one.
Encoding each heap in two bits uses only constants and wiring, and the low bit of the replacement heap is the input parity. This gives a constant-depth, polynomial-size reduction to winning-move selection, although every encoded position is winning.
\end{proof}

\subsection{First bypass: two-frame restoration}
\label{sec:multiframe}

The fundamental idea is to avoid computing the \emph{global} nim-sum from scratch. Instead, we exploit that each move changes only one heap, so the \emph{nimber difference} between consecutive positions is locally computable.

\begin{definition}[Nimber difference]
\label{def:nimber-difference}
Given consecutive positions $P$ and $P'$ in a play sequence, define the \emph{nimber difference}
\[
\Delta(P,P') \;=\; \operatorname{nim}(P)\ \oplus\ \operatorname{nim}(P').
\]
\end{definition}

In NIM, if $P'$ is obtained from $P$ by changing heap $i$ from $x_i$ to $x_i'$, then
\[
\Delta(P,P') \;=\; x_i \oplus x_i',
\]
a purely local computation on the changed heap.

\begin{lemma}[Bounded-change differences]
\label{lem:nimber-diff-ac0}
Fix a constant $t\in\N$.
Let $P$ and $P'$ be NIM positions whose heap vectors differ in at most $t$ heaps.
Then $\Delta(P,P')$ can be computed by an $\AC^0$ circuit of polynomial size (in the input length).
\end{lemma}

\begin{proof}
See Appendix~\ref{app:nimberdiff}.
\end{proof}

Membership in $\AC^0$ does not guarantee a small or readily learnable network. For the complete two-frame move selector built on this difference computation, our modular construction gives an upper bound of 42 threshold layers; one shootout-controller step has a bound of 70 (Appendix~\ref{app:impl}). These are constructive bounds, not necessary depths. Any Boolean function has a depth-two AND--OR representation with negated inputs, but its size can be exponential; $\AC^0$ requires polynomial size as the input grows. We have not optimised this depth--size trade-off. Removing the expressivity obstruction therefore leaves practical network design, training procedures and hyperparameter choices open. The layer bounds do not establish the cause of the experimental learning plateaus.

\subsubsection{Two-frame local restoration rule}
\label{sec:restoration}

The next observation is that if, after our move, we make the opponent face a zero nim-sum position, then the opponent's next move produces some nimber difference $\delta$. If we respond with a move that produces the \emph{same} nimber difference $\delta$, we restore nim-sum zero.

\begin{proposition}[Local restoration principle]
\label{prop:restore}
Suppose the \emph{opponent} moves from a position $Q$ with $\operatorname{nim}(Q)=0$ to a position $Q'$ (so it is now our turn).
Let $\Delta(Q,Q')=\operatorname{nim}(Q)\oplus \operatorname{nim}(Q')$ denote the nimber difference caused by the opponent's move.
If we respond with a move $Q' \to R$ such that
\[
\Delta(Q',R) \;=\; \Delta(Q,Q'),
\]
then $\operatorname{nim}(R)=0$.
\end{proposition}

\begin{proof}
By definition, $\operatorname{nim}(Q') = \operatorname{nim}(Q) \oplus \Delta(Q,Q')$.
Since $\operatorname{nim}(Q)=0$ (by hypothesis), we have $\operatorname{nim}(Q') = \Delta(Q,Q')$.
The player's move creates a new position $R$ with $\operatorname{nim}(R) = \operatorname{nim}(Q') \oplus \Delta(Q',R)$.
Substituting the condition $\Delta(Q',R) = \Delta(Q,Q')$:
\[
\operatorname{nim}(R) \;=\; \operatorname{nim}(Q') \oplus \operatorname{nim}(Q') \;=\; 0.
\]
Thus, matching the opponent's difference restores the invariant.
\end{proof}

Moreover, a concrete restoration move can be selected in $\AC^0$ from $(Q,Q')$ by computing $\delta=\Delta(Q,Q')$ and then choosing (with a fixed tie-break rule) a heap $i$ such that $x_i' = x_i \oplus \delta < x_i$ (Appendix~\ref{app:impl}).

\begin{figure}[t]
\centering
\begin{tikzpicture}[
    node distance=1.5cm,
    auto,
    font=\small,
    p_pos/.style={
        rectangle,
        draw=green!50!black,
        fill=green!5,
        thick,
        rounded corners,
        align=center,
        text width=2.8cm,
        minimum height=1.6cm,
        inner sep=4pt
    },
    n_pos/.style={
        rectangle,
        draw=red!50!black,
        fill=red!5,
        thick,
        rounded corners,
        align=center,
        text width=2.8cm,
        minimum height=1.6cm,
        inner sep=4pt
    },
    arr/.style={-{Latex[length=2.5mm]}, thick, black!70},
    restore_arr/.style={-{Latex[length=2.5mm]}, very thick, blue!80}
]
    \node[p_pos] (Q) {
        \textbf{Start ($Q$)}\\
        Heaps: $\{3, 5, 6\}$\\
        \textcolor{green!40!black}{\textbf{nim-sum} $\mathbf{= 0}$}
    };

    \node[n_pos, right=2.8cm of Q] (Qp) {
        \textbf{After Opponent ($Q'$)}\\
        Heaps: $\{3, \mathbf{4}, 6\}$\\
        \textcolor{red!40!black}{\textbf{nim-sum} $\mathbf{= 1}$}
    };

    \node[p_pos, right=2.8cm of Qp] (R) {
        \textbf{Restored ($R$)}\\
        Heaps: $\{\mathbf{2}, 4, 6\}$\\
        \textcolor{green!40!black}{\textbf{nim-sum} $\mathbf{= 0}$}
    };

    \draw[arr] (Q) -- node[align=center, above, font=\footnotesize] {Opponent moves\\$5 \to 4$}
                      node[align=center, below, font=\footnotesize] {$\delta = 5 \oplus 4 = \mathbf{1}$} (Qp);

    \draw[restore_arr] (Qp) -- node[align=center, above, font=\footnotesize, text=blue!80] {\textbf{Agent restores}\\$3 \to 2$}
                               node[align=center, below, font=\footnotesize, text=blue!80] {Matches $\delta=\mathbf{1}$\\$(3 \oplus 2 = 1)$} (R);

\end{tikzpicture}
\caption{A worked restoration example (three heaps).
Starting from a $P$-position $Q$ (green), the opponent changes heap 2 ($5\to 4$), creating a nimber difference $\delta=1$.
The agent instantly restores the $P$-position by changing heap 1 ($3\to 2$), which imposes the matching difference $3\oplus 2 = 1$.}
\label{fig:restoration-example}
\end{figure}

\paragraph{Interpretation (``mirror the delta'').}
Once we have made the opponent face a $P$-position, we do not need to recompute $\operatorname{nim}(\cdot)$ from scratch. We only need to mirror the observed change $\Delta(Q,Q')$ in our response move.
Figure~\ref{fig:restoration-example} illustrates this mechanism on a concrete three-heap instance.

Proposition~\ref{prop:restore} is structural: it describes a response pattern that keeps returning the game to nim-sum zero, provided the player can (i) compute $\delta=\Delta(Q,Q')$ from the last two frames and (ii) choose a legal move $Q'\to R$ that realises the same $\delta$. In NIM, when the opponent changes exactly one heap, $\delta$ is simply $x_i\oplus x_i'$, and testing whether a candidate response move realises $\delta$ is a local check on the heap being changed.

The significance for learnability is representational. A two-frame policy network can condition its response on $(Q,Q')$ and hence on the locally computable $\delta$, while a single-frame policy must reconstruct the relevant global information from $Q'$ alone. The experiments in Section~\ref{sec:experiments} examine this contrast through supervised restoration learning and imitation-trained rollout policies. The reported networks are policy-only; these results do not establish that self-play discovers the constructions.

\subsubsection{Implementation overview}
An $\AC^0$ implementation of the two-frame delta and restoration components can be organised into:
\begin{itemize}[leftmargin=*]
\item a binary encoding scheme for heap vectors (Appendix~\ref{app:impl}),
\item a constant-depth detection of which heaps changed between frames,
\item an $\AC^0$ circuit computing $\Delta$ on those changed heaps (Lemma~\ref{lem:nimber-diff-ac0}),
\item an $\AC^0$ selection mechanism for a response move realising a target $\delta$.
\end{itemize}

\subsection{Second bypass: structured rollout policies}
\label{sec:search-gap}

Theorem~\ref{thm:singleframe-impossible} shows that no $\AC^0$ circuit can evaluate the nim-sum in a single forward pass.
A natural follow-up is whether \emph{planning}---testing candidate moves and simulating play, as in Monte-Carlo Tree Search (MCTS)---can compensate.

We analyse a stylised rollout interface: a planning algorithm queries a \emph{rollout policy} $\pi$ by simulating a complete game in which \emph{both} players follow $\pi$ from a start position, and it observes only the terminal win/loss bit.
This abstraction isolates the information that a weak policy can expose through repeated simulation; it is not a claim about the optimisation dynamics of AlphaZero-style training (cf.\ Section~\ref{sec:scope}).

\begin{definition}[Rollout policy and rollout outcome]
\label{def:rollout-policy}
A deterministic rollout policy maps a nonterminal $NB$-bit NIM position to a legal move, encoded by a heap index and its new value; on the terminal position it outputs a halt symbol. It is $\AC^0$-bounded when this map has polynomial-size, constant-depth circuits.
A rollout has both players follow that policy until termination. Its outcome $\mathrm{RO}_{\pi}(P)$ is 1 if the first player wins and 0 otherwise, including at a terminal start. The interface returns only this terminal bit, not a trajectory or move count.
\end{definition}

\begin{definition}[Restricted rollout measurements and readout]
\label{def:agent-types}
Fix a root position $P$. An admissible query starts at $P$ or a position obtained by changing one heap of $P$; the changed index and old/new values are known to the readout. A legal one-move successor is an admissible query. Arbitrary transformations of the whole root are excluded.
For the single-channel lower bound, one fixes a policy $\pi_j$ from Proposition~\ref{prop:rollout-sufficiency}, independently of $P$. Writing $\beta=\mathrm{RO}_{\pi_j}(P)$, the agent's final move must have the form $F_N(P,\beta)$, where $F_N$ is a polynomial-size, constant-depth Boolean circuit family. Any value head, query corrections or other processing must be included in this readout bound. The number of queries alone does not impose that bound.
A multi-channel construction may query distinct members of the same structured policy family and apply a polynomial-size, constant-depth readout to $P$ and their outcome bits. A \emph{policy head} means a rollout policy, not simply a parallel output neuron. We count simulated game moves separately from the circuit work per move.
\end{definition}

Allowing arbitrary start positions would defeat the single-channel restriction: querying $\pi_0$ on heaps consisting of the $k$th bits of $P$ returns $\sigma_k(P)$. Likewise, an unrestricted sequential controller could compute the nim-sum itself. The lower bound therefore concerns the stated measurement/readout model, not general MCTS or all single-policy algorithms.

\begin{remark}[Why multiple heads are not redundant]
\label{rem:why-multi-head}
Since $\AC^0$ is closed under parallel composition, one could compute all $k$ policies simultaneously as a single $\AC^0$ circuit.
The distinction is not about computational power at the circuit level but about the \emph{rollout interface}: each rollout query simulates a complete game under \emph{one} policy, and at each game step only that policy determines the next move.
A single rollout trajectory under a fixed policy $\pi$ produces one outcome bit; merging $k$ circuits into a single circuit does not change the fact that any one rollout follows a single deterministic rule, yielding a single bit of information.
Multiple heads are useful because the agent can issue \emph{separate} rollout queries under different policies $\pi_j$, producing different game trajectories that reveal \emph{independent} linear functionals of the nim-sum (cf.\ Propositions~\ref{prop:rollout-sufficiency} and~\ref{prop:single-policy-bound}).
\end{remark}

\paragraph{Finite-state shootout controller (FSM over policy heads).}
\label{par:shootout-controller}
The multi-channel classifier can stop early after a nonzero channel outcome, without changing its Boolean decision rule (the OR of all channel bits). This limited scheduling optimisation does not give the single-channel readout an unrestricted sequential controller.
To make this adaptivity explicit (without changing the $\AC^0$ constraints on each head), it is convenient to view head-selection as a small \emph{finite-state controller} that sits outside the heads and only sees the terminal win/loss bit of each rollout.

Formally, fix rollout heads $\pi_0,\dots,\pi_{k-1}$.
A \emph{finite-state shootout controller} is a tuple
\[
\mathcal{C} = (\mathcal{S}, s_{\mathrm{init}}, g, \delta, \mathsf{halt}),
\]
where $\mathcal{S}$ is a finite set of controller states, $s_{\mathrm{init}}\in\mathcal{S}$ is the initial state,
$g\colon \mathcal{S}\to\{0,\dots,k-1\}$ selects which head to use for the next rollout query,
$\delta\colon \mathcal{S}\times\{0,1\}\to\mathcal{S}$ updates the controller state after observing a rollout outcome bit
$r=\mathrm{RO}_{\pi_{g(s)}}(P)$, and $\mathsf{halt}\subseteq \mathcal{S}$ is a set of halting states associated with an output decision.
(When $\mathsf{halt}=\emptyset$, the controller simply runs for a fixed number of queries and outputs a decision at the end.)

This viewpoint does \emph{not} increase the per-step computational power of the rollout heads: each $\pi_j$ remains $\AC^0$-bounded.
Instead, the controller organises \emph{which} $\AC^0$ head is iterated in a shootout and \emph{when} to stop querying.
Figure~\ref{fig:fsm-shootout-controller} depicts a particularly simple controller that turns the $B$ structured rollout heads from Proposition~\ref{prop:rollout-sufficiency} into a sequential ``bit-query'' procedure for deciding whether $\sigma(P)=\operatorname{nim}(P)$ is zero (and hence whether $P$ is a $P$-position).

The checked implementation includes the outer controller: it selects a head, executes its rollout, records the outcome and either reloads the original board for the next head or halts. For classification it stops at the first outcome~1. For move selection our implementation runs all $B$ heads, since an early classification decision need not supply the invariant bits required by the move readout. Each controller tick is itself an $\AC^0$ map. The original board remains available as a read-only input throughout; it is not an oracle for its nim-sum. Appendix~\ref{app:impl} specifies the storage and update costs.

\begin{figure}[t]
\centering
\begin{tikzpicture}[>=Latex, node distance=16mm and 18mm, font=\small]
\tikzset{
  state/.style={draw, rounded corners, minimum height=8mm, minimum width=20mm, align=center},
  term/.style={draw, rounded corners, thick, minimum height=8mm, minimum width=28mm, align=center},
  arrow/.style={->, thick}
}

\node[state] (q0) {$q_0$\\ use $\pi_0$};
\node[state, right=of q0] (q1) {$q_1$\\ use $\pi_1$};
\node[inner sep=0pt, right=of q1] (dots) {\Large$\cdots$};
\node[state, right=of dots] (qB) {$q_{B-1}$\\ use $\pi_{B-1}$};

\node[term, above=12mm of dots] (nonzero) {$\sigma(P)\neq 0$\\ (declare $N$)};
\node[term, below=12mm of qB] (zero) {$\sigma(P)=0$\\ (declare $P$)};

\draw[arrow] (q0) -- node[below] {$r_0=0$} (q1);
\draw[arrow] (q0) to[bend left=25] node[left] {$r_0=1$} (nonzero);

\draw[arrow] (q1) -- node[below] {$r_1=0$} (dots);
\draw[arrow] (q1) to[bend left=18] node[above left] {$r_1=1$} (nonzero);

\draw[arrow] (dots) -- node[below] {$r_j=0$} (qB);
\draw[arrow, dashed] (dots) to[bend left=15] node[above] {$\exists j:\ r_j=1$} (nonzero);

\draw[arrow] (qB) to[bend left=15] node[above right] {$r_{B-1}=1$} (nonzero);
\draw[arrow] (qB) -- node[right] {$r_{B-1}=0$} (zero);

\end{tikzpicture}
\caption{A simple finite-state shootout controller for deciding whether $\sigma(P)=\operatorname{nim}(P)$ is zero using the $B$ rollout heads of Proposition~\ref{prop:rollout-sufficiency}.
The controller queries heads sequentially and halts as soon as it can certify $\sigma(P)\neq 0$.
Reaching the final transition to $\sigma(P)=0$ implicitly means $r_0=\cdots=r_{B-1}=0$.
}
\label{fig:fsm-shootout-controller}
\end{figure}

\noindent\textbf{Integration (channels vs.\ protocol).}
It is helpful to reinterpret Part~(ii) of Theorem~\ref{thm:search-gap} in an ``information-channel'' language: each policy head $\pi_j$ defines a distinct rollout \emph{measurement channel} which, when iterated to termination, returns a single bit $r_j=\mathrm{RO}_{\pi_j}(P)$ that depends on the hidden $B$-bit invariant $\sigma(P)$ (cf.\ the identities in Proposition~\ref{prop:rollout-sufficiency}).
With a \emph{single} head, repeated deterministic root rollouts return the \emph{same} linear functional $\ell(\sigma)$; admissible one-heap modifications give its known affine corrections (Proposition~\ref{prop:single-policy-bound}). These measurements supply no further unknown invariant bit.
With \emph{multiple} heads, the agent has access to linearly independent invariant measurements and can reconstruct $\sigma(P)$ from their outcomes (Proposition~\ref{prop:rollout-sufficiency}).
The finite-state shootout controller introduced above is the natural generalisation of this idea: it does not change what any head can compute, but organises the heads into an \emph{adaptive querying protocol}---the controller decides which channel to query next (and when to stop, or when to re-query for robustness) based on previously observed rollout outcomes.
In this view, multi-head provides the \emph{channels}, while the FSM provides the \emph{sequential control} that turns those channels into a practical evaluation procedure.

\paragraph{Depth amplification.}
Although a single $\AC^0$ circuit cannot compute parity, iterating an $\AC^0$ policy for $L$ game steps yields an effective depth $\Theta(L)$ computation, which can reveal parity information.
The next proposition makes this explicit by exhibiting $B$ simple $\AC^0$ rollout heads whose outcomes recover all $B$ bits of the nim-sum.

\begin{proposition}[Structured rollout sufficiency]
\label{prop:rollout-sufficiency}
The $B$ structured rollout policies below, followed by an $\AC^0$ move selector, achieve strong mastery of NIM on $N$ heaps of bit-width $B$. They use $B$ root rollouts and $O(BN2^B)$ simulated game moves per decision. Each policy update and the final readout have polynomial circuit size in $NB$. For $B=O(\log N)$ the resulting bit work is polynomial in $N$.
\end{proposition}

\begin{proof}[Proof outline]
For each $j\in\{0,\dots,B-1\}$ define a deterministic rollout policy $\pi_j$ that repeatedly subtracts $2^j$ from the leftmost heap of size at least $2^j$ (and otherwise subtracts $1$ from the leftmost nonzero heap). Both simulated players follow this policy; the actual opponent may choose any legal move.
A rollout under $\pi_j$ has a well-defined total number of moves whose parity is
\[
\mathrm{RO}_{\pi_0}(P)=\sigma_0,\qquad
\mathrm{RO}_{\pi_j}(P)=\sigma_j\oplus\sigma_0\quad(j\ge 1),
\]
where $\sigma=\operatorname{nim}(P)$ and $\sigma_j$ denotes its $j$th bit.
The search controller first queries $\pi_0$ to obtain $\sigma_0$, then queries each $\pi_j$ ($j\ge 1$) to obtain $\sigma_j\oplus\sigma_0$, and recovers $\sigma_j = (\sigma_j\oplus\sigma_0)\oplus\sigma_0$; thus $B$ rollouts under $B$ distinct heads recover all $B$ bits of~$\sigma$.
Given $\sigma\ne0$, select the first heap $h_i$ with $h_i\oplus\sigma<h_i$ and reduce it to $h_i\oplus\sigma$. If $\sigma=0$, take any legal move, or halt at the terminal position. No successor enumeration is needed.
A complete proof is in Appendix~\ref{app:rollout-proofs}.
\end{proof}

\paragraph{A single policy head yields only one bit.}
The preceding result relies on having \emph{multiple} rollout channels.
For one fixed member of this structured policy family, admissible queries reveal known affine functions of the same invariant bit. This statement does not concern arbitrary rollout policies.

\begin{proposition}[Single-policy information bound]
\label{prop:single-policy-bound}
Fix $j$ and the rollout policy $\pi_j$ from the proof outline of Proposition~\ref{prop:rollout-sufficiency}.
For any position $P$, the rollout outcome $\mathrm{RO}_{\pi_j}(P)$ is a fixed linear functional $\ell_j(\sigma(P))$ of the nim-sum bits (indeed, $\ell_0(\sigma)=\sigma_0$ and $\ell_j(\sigma)=\sigma_j\oplus\sigma_0$ for $j\ge 1$).
Moreover, if $P_i$ is obtained from $P$ by modifying one heap (as when evaluating candidate moves), then
\[
\mathrm{RO}_{\pi_j}(P_i)=\ell_j(\sigma(P))\oplus c_i,
\]
where the correction bit $c_i$ is computable from the known heap values and the chosen modification.
Hence root and one-heap-modification queries to this fixed head reveal only known affine functions of $\ell_j(\sigma(P))$.
\end{proposition}

\begin{proof}
The linearity claim follows from the explicit formulas above.
If $P_i$ differs from $P$ by changing one heap from $h$ to $h'$, then $\sigma(P_i)=\sigma(P)\oplus(h\oplus h')$ and $\ell_j(\sigma(P_i))=\ell_j(\sigma(P))\oplus \ell_j(h\oplus h')$, where $\ell_j(h\oplus h')$ is known from the candidate move.
\end{proof}

\begin{remark}[An illustrative independent outcome-flip model]
\label{rem:noisy-rollout-decay}
Suppose an ideal terminal bit $Y$ is corrupted as $\widetilde Y=Y\oplus E_1\oplus\cdots\oplus E_L$, where the $E_i$ are independent Bernoulli variables of parameter $0\le\varepsilon<1/2$. Then
\[
\Pr[\widetilde Y=Y]=\frac{1+(1-2\varepsilon)^L}{2}.
\]
The formula follows by multiplying the expectations of $(-1)^{E_i}$. With fresh independent flips for each repetition, the least number of observations needed to distinguish the two possible ideal bits with worst-case error at most $1/4$ grows as $a^{-2}$, where $a=(1-2\varepsilon)^L$. Appendix~\ref{app:noise-samples} states and proves explicit bounds over real parameters. Repeating one corrupted observation with the same randomness supplies no additional information.
At $\varepsilon=0.01$ and $L=50$, agreement is approximately $0.682$, so this example is not yet close to a fair coin.
This is a channel-noise illustration, not a theorem about erroneous game moves: a move error need not flip the terminal bit, and can change later states and rollout length. Correlated flips can cancel in pairs, eliminating the claimed decay. Exact structured rollouts have no such noise.
\end{remark}

\paragraph{The initialisation boundary for restoration.}
Matching two successive differences yields $\operatorname{nim}(R)=\operatorname{nim}(Q)$, whether or not that value is zero. Thus Proposition~\ref{prop:restore} does not supply a classifier for an arbitrary starting position.
For example, complete a matching-delta policy by subtracting one from the first nonempty heap whenever no matching response exists or no previous frame is available. Starting from $(2)$, it plays $(2)\to(1)\to(0)$ and the first player loses, although $(2)$ is winning. The fallback after the first move is needed because its delta is $2\oplus1=3$, which cannot be realised from $(1)$.
This counterexample does not rule out every possible verifier construction; it shows that conditional restoration and an unspecified fallback do not establish one. We make no universal-verifier claim. Initial winning-move selection can instead use the multi-channel construction above or the growing recurrent scan in Section~\ref{sec:recurrent-bypass}.

\begin{figure}[t]
\centering
\begin{tabular}{>{\raggedright\arraybackslash}p{0.28\linewidth}>{\raggedright\arraybackslash}p{0.29\linewidth}>{\raggedright\arraybackslash}p{0.31\linewidth}}
\hline
\textbf{Resource} & \textbf{What stays in $\AC^0$} & \textbf{Additional information or time} \\
\hline
One structured channel & Policy step and final readout & One complete rollout; one invariant bit \\
$B$ structured channels & Each policy step; reconstruction and move selection & $B$ complete rollouts supply all invariant bits \\
Two-frame transition & Local difference and binary move selection & Previous zero-nim-sum board and a legal transition \\
Persistent row scan & One state update and its output bits & Two scans per decision; $O(B+\log N)$ state bits \\
\hline
\end{tabular}
\caption{The three bypasses use different resources. Conditional restoration does not classify arbitrary starting positions; the rollout and recurrent constructions supply additional sequential computation.}
\label{fig:rollout_gap}
\end{figure}

\begin{theorem}[Restricted bound and bypasses]
\label{thm:search-gap}
The following statements hold under Definition~\ref{def:agent-types}.
\begin{enumerate}
\item[\textup{(i)}] For each fixed $B\ge2$, let $P$ be uniform on $\{0,\ldots,2^B-1\}^N$ conditional on $\sigma(P)\ne0$. For one fixed structured head $\pi_j$ and an $\AC^0$ readout $F_N(P,\mathrm{RO}_{\pi_j}(P))$, winning-move success is at most
\[
\frac{2}{2^B-1}+\operatorname{negl}(N).
\]
Here $N\ge1$, and a negligible error is a nonnegative sequence smaller than every inverse polynomial for sufficiently large $N$, with circuit depth, polynomial size exponent and $B$ fixed. More precisely, a uniform parity-correlation bound $\eta_N$ for the fixed-bit readout indicators gives the finite upper bound $2/(2^B-1)+(2^B-1)\eta_N$; the indicators and correlation premise are specified in Appendix~\ref{app:search-gap-proof}. The leading constant is asymptotically attainable by guessing one nonzero nim-sum in each channel fibre.
\item[\textup{(ii)}] For any $B$, the $B$ structured heads and an $\AC^0$ readout select an optimal move using $O(BN2^B)$ rollout steps, as in Proposition~\ref{prop:rollout-sufficiency}.
\item[\textup{(iii)}] Given a legal transition from a known zero-nim-sum predecessor, a two-frame $\AC^0$ selector restores zero. Once a player has established that invariant, it maintains it against every legal opponent response. This statement does not supply the initial winning move from an arbitrary position.
\end{enumerate}
\end{theorem}

\begin{proof}[Proof outline]
For part (i), fixing the observed bit leaves an $\AC^0$ move selector. The selected heap's old and new values determine a proposed nim-sum. Expanding each exact nim-sum event into its finitely many parity characters and applying the $\AC^0$ parity-correlation bound gives the displayed estimate; Appendix~\ref{app:search-gap-proof} gives the full reduction and conditional-probability calculation. Parts (ii) and (iii) follow from Propositions~\ref{prop:rollout-sufficiency} and~\ref{prop:restore}, with the selector described in Appendix~\ref{app:impl}.
\end{proof}


\subsection{Third bypass: persistent row-wise recurrence}
\label{sec:recurrent-bypass}

The two-frame construction supplies the relevant history explicitly, while the multi-head construction obtains additional information through complete rollouts. A third possibility is to let the agent build and retain its own state by processing the board sequentially. The recurrent update remains simple, but the number of updates is allowed to grow with the input. The construction uses the elementary running-XOR algorithm. In correspondence reproduced by Friedman \citep{friedman2017deeplearning}, John Schulman proposed feeding NIM heaps one at a time to a recurrent network and reported preliminary supervised nim-sum experiments. Here we make the shallow updates, separator and state protocols, and move-selection constructions explicit, distinguishing their representational guarantees from learning through self-play.

\paragraph{Serialising positions and retaining state.}
Write position $P_t=(x_{t,1},\ldots,x_{t,N})$, with each heap represented by one $B$-bit row. Feed successive positions as the stream
\[
\underbrace{x_{1,1},\ldots,x_{1,N}}_{P_1},\ \mathtt{SEP},\quad
\underbrace{x_{2,1},\ldots,x_{2,N}}_{P_2},\ \mathtt{SEP},\quad\ldots .
\]
Each row arrives as a block of $B$ bits together with a row/separator tag. A separator marks the end of a complete position; the agent chooses a move only after that position has been read. The next observed position is then appended to the stream. For comparisons of individual heaps, indices remain aligned across positions, including rows for empty heaps.
A scan of the current board suffices for nim-sum evaluation without earlier game positions; cross-position memory additionally supports the comparisons below. The hidden state is carried across rows and position boundaries. A separator resets the accumulator for the next scan while preserving selected information about the completed position; it does not reset the entire memory.

\begin{proposition}[Sequential nim-sum accumulation]
\label{prop:streaming-nimsum}
There is a recurrent Boolean update of constant depth and size $O(B)$, implementable with fixed-scale quantised threshold gates, that computes the nim-sum of an $N$-heap position after $N$ row updates. With $O(B)$ state bits, it can also retain the previous completed nim-sum and report the XOR difference between consecutive completed positions.
\end{proposition}
\begin{proof}
Initialise a $B$-bit accumulator to zero and update it by
\[
s_{t,0}=0,\qquad s_{t,i}=s_{t,i-1}\oplus x_{t,i}\quad (1\le i\le N).
\]
Induction on $i$ gives $s_{t,i}=\bigoplus_{j=1}^{i}x_{t,j}$, so the final accumulator is $\sigma_t:=\operatorname{nim}(P_t)$.
Each output bit of an update is the XOR of two bits, which has a constant-size, constant-depth AND/OR/NOT implementation. These gates can be realised with fixed-scale threshold gates and input negations, with constants independent of $N$ and $B$.
The summary state is $(s,q,v)$: two $B$-bit registers and one validity bit, initialised to $(0,0,0)$. A row $x$ updates it to $(s\oplus x,q,v)$. At a separator, output $s$ and, when $v=1$, the difference $q\oplus s$; then update the state to $(0,s,1)$. The first separator therefore produces a summary but no previous-position difference. Induction over completed positions shows that $q$ stores exactly the preceding summary and $s$ is the XOR of the rows since the last separator. The summary mechanism uses $2B+1$ state bits; a reset between games restores $(0,0,0)$. All assignments have constant depth and size $O(B)$.
\end{proof}

\paragraph{From a summary to a move.}
Computing $\sigma_t$ does not by itself retain the heap rows needed to choose a move. One implementation replays the current position after the first scan. If $\sigma_t\ne0$, select the first row $i$ for which
\[
x_{t,i}'=x_{t,i}\oplus\sigma_t<x_{t,i}.
\]
Such a row exists: choose a heap containing the highest set bit of $\sigma_t$. Replacing that heap by $x_{t,i}'$ leaves nim-sum zero. XOR, binary comparison and conditional selection are constant-depth operations; the replay uses $O(B+\log N)$ state bits to retain the summary, a selected move and its row index. The explicit protocol is
\[
\mathtt{rows}(P_t),\ \mathtt{SEP},\ \mathtt{rows}(P_t),\ \mathtt{SEP}.
\]
The first separator stores the summary and switches to replay; the second emits the move and returns to scan mode. This takes exactly $2N+2$ updates, with one additional reset token at the start of a game. If $\sigma_t=0$, choose the first nonempty heap and decrement it, or report termination when all heaps are empty. Lean verifies the complete replay controller, including its binary move output and its fixed-scale threshold implementation. With an $r$-bit counter satisfying $N<2^r$, its state uses $4+5B+3r$ bits. The strict inequality allows the counter to hold the end value $N$, whereas an output index alone only needs $N\le2^r$.
Alternatively, store the current board in $O(NB)$ bits during the initial scan and use a constant-depth selector at the separator, as in Appendix~\ref{app:impl}. This avoids replay at the cost of board-sized memory. The distinction matters for a stream in which each position is supplied only once. Both variants have checked bit-level circuits, fixed-scale network implementations and winning-play guarantees. The board-storage variant uses $NB+B+r$ state bits and $N+1$ updates per decision, with $N<2^r$, plus a reset update when initialising the controller.

\paragraph{Connecting recurrence to the history bypass.}
The retained summaries allow the agent to compare consecutive positions through
\[
\sigma_{t-1}\oplus\sigma_t
=\operatorname{nim}(P_{t-1})\oplus\operatorname{nim}(P_t)
=\Delta(P_{t-1},P_t).
\]
When one heap changes from $a$ to $b$, this equals $a\oplus b$. If the preceding position has nim-sum zero, the difference is exactly the invariant that the restoration rule must cancel. Storing the preceding board additionally permits direct row-by-row comparison as the new board arrives. A summary of $O(B)$ bits preserves the invariant, rather than the full game history; arbitrary access to old positions requires additional storage or replay. The difference always refers to consecutive completed boards actually supplied to the controller. If these are only the boards at the player's decision times, they usually span two game moves. Comparing just the opponent's move instead requires the player's own post-move board as the preceding frame.
The initial scan also computes the invariant from an arbitrary starting position, without assuming an earlier zero-nim-sum frame. Thus memory across row updates is sufficient for this route; an earlier game position is useful for transition comparisons but is not required to compute the current nim-sum. This initialisation uses the growing recurrent horizon and is therefore an additional resource beyond the fixed-latency two-frame construction.

\paragraph{Why the lower bound is respected.}
Unrolling $N$ constant-depth updates gives a computation of depth $O(N)$, so Proposition~\ref{prop:streaming-nimsum} lies outside the fixed-$T$ premise of Theorem~\ref{thm:sim-ac0}. A one-step recurrent dependency can nevertheless carry information accumulated over the whole stream. Merely appending positions and separators to a fixed-depth input does not supply this growing computation horizon. The bypass comes from sequential computation with retained state, not from the delimiter itself.
For growing heap bit-width, the accumulator also needs $O(B)$ bits; a fixed-width hidden state is not claimed to store arbitrary-size invariants or histories. The construction establishes representational sufficiency for NIM, but does not establish that training a recurrent network will discover the update or preserve it reliably over long games. We do not report recurrent training experiments here.

\subsection{What the three bypasses establish}
\label{sec:bypass-scope}

\begin{proposition}[Shallow components and complete strategies]
\label{prop:bypass-interfaces}
For $N\ge1$ heaps of $B\ge1$ bits, the following constructions use polynomial-size, constant-depth Boolean components, realisable by literal-threshold networks with weights and thresholds in $\{0,1\}$.
\begin{enumerate}[leftmargin=*,label=(\roman*)]
\item The two-frame selector maps a legal transition $Q\to Q'$ from $\operatorname{nim}(Q)=0$ to a legal binary-encoded response restoring zero. Retaining the response board as the next predecessor gives a winning strategy from this established invariant, with the opponent to move.
\item Each structured rollout policy and the readout from the board and its $B$ outcome bits are shallow. Executing the rollouts and then the readout gives a winning strategy from any nonzero nim-sum board, using at most $BN(2^B-1)$ simulated game moves per decision.
\item The recurrent state update is shallow. Iterating it over the replay protocol gives a winning strategy from any nonzero nim-sum board, with $O(B+\log N)$ memory and $2N+2$ updates per decision.
\end{enumerate}
Here ``winning strategy'' quantifies over every legal opponent response. Polynomial size and constant depth refer to the indicated component, with depth independent of $N$ and $B$; they do not bound the number of sequential component evaluations.
\end{proposition}

\begin{proof}[Proof outline]
Appendix~\ref{app:impl} gives explicit candidate tests, first-index encoding and a fixed-scale implementation. The two-frame circuit computes the changed heap's XOR difference, which equals the current nim-sum under the stated promise. The multi-head readout reconstructs that same invariant from the executed rollout outcomes. The recurrent controller obtains it by accumulation and uses the stored result during replay. In each case the selected move restores zero; every legal opponent move makes the nim-sum nonzero, and every move decreases the total number of stones. This proves termination and the winning guarantee. For the two-frame strategy, the newly restored board is precisely the frame retained for the next response. For recurrence, the finite controller state persists between decisions. The rollout and recurrent step bounds are those proved above.
\end{proof}

These are three ways around the \emph{same restricted inference problem}, not three shallow algorithms for unrestricted parity. The two-frame construction changes the input and imposes a promise. The other constructions allow processing time to grow. Their complete board-to-winning-move maps cannot themselves form an $\AC^0$ family: otherwise applying either to $(2,b_1,\ldots,b_m)$ would contradict Theorem~\ref{thm:singleframe-impossible}. Parallel output heads alone do not supply the rollout computation, and separators alone do not supply a recurrent scan. Appendix~\ref{app:impl} illustrates the explicit network constructions and explains their conservative depth and layer bounds.

\section{Learning impartial games with neural networks}
\label{sec:learnability}

The results above are primarily \emph{representational}: they identify what information a bounded-depth, fixed-scale quantised model can or cannot compute from its input, and how history, multiple rollout heads, or a growing recurrent horizon can supply the missing information.
This section connects the theory to the training protocols used in Section~\ref{sec:experiments} and highlights the main modelling choice ($\AC^0$ vs.\ $\TC^0$).

\subsection{Practical takeaways}
\begin{itemize}[leftmargin=*]
\item \textbf{Single-frame is brittle to parity.}
In the worst case, fixed-scale quantised bounded-depth inference cannot reliably represent the nim-sum or a winning move selector (Section~\ref{sec:impossibility}).

\item \textbf{Three bypasses with different resources.}
A two-frame input exposes the transition signal $\Delta(P,P')$ and supports restoration (Section~\ref{sec:restoration}); multiple rollout heads recover the missing nim-sum bits through depth amplification (Section~\ref{sec:search-gap}); persistent recurrence accumulates the invariant over sequential row updates and retains summaries across positions (Section~\ref{sec:recurrent-bypass}). The last route trades fixed total latency for a growing processing horizon.

\item \textbf{Optimisation remains nontrivial.}
Even when a solution is expressible in $\AC^0$, discovering it from sparse win/loss rewards can be difficult; our experiments therefore isolate a dense restoration supervision signal and separately report rollout-based classification.
\end{itemize}

\subsection{Teacher-guided data as a controlled source of transitions}
\label{sec:teacher}

Let $\pi_{\mathrm{opt}}$ denote an optimal NIM policy (move to nim-sum~$0$ whenever possible).
For a noise parameter $\varepsilon\in[0,1]$, define a noisy teacher $\pi_T^{\varepsilon}$ that plays according to $\pi_{\mathrm{opt}}$ with probability $1-\varepsilon$ and otherwise chooses a uniformly random legal move.
Occasional random moves broaden state coverage and, crucially for two-frame agents, generate diverse transition pairs $(Q,Q')$ from which the local difference signal $\Delta(Q,Q')$ can be inferred.

\subsection{Approximate majority, parity, and the \texorpdfstring{$\AC^0$}{AC0} abstraction}
\label{sec:majority-vs-parity}

One might object that modern networks can implement majority-like decisions, suggesting $\TC^0$ (threshold circuits) as the ``right'' complexity class.
Indeed, if one allows scale-adaptive thresholds/biases or fan-in dependent re-scaling, constant-depth threshold models can represent parity-type computations.

Our focus is a restricted literal-threshold regime: fixed total depth, nonnegative weights on Boolean literals, and thresholds bounded in units of the smallest nonzero weight.
In this regime, each gate has a constant-size witness and collapses to a constant-width DNF, yielding an $\AC^0$ abstraction (Theorem~\ref{thm:sim-ac0}).
Constant-depth $\AC^0$ circuits can implement approximate counting \citep{ajtai1993approxcount}, yet they have exponentially small correlation with parity \citep{furst1984parity,hastad1986thesis,smolensky1987algebraic,razborov1987lower}.
Learning parity depends on the architecture and training setting; both difficulties and positive results have been studied \citep{thornton1996parity,thornton1996bp,daniely2020parities}. The representational lower bound here applies only to the explicit gate model.
Section~\ref{sec:spectral-view} gives a complementary Fourier/spectral interpretation of this local-versus-global parity separation.

\subsection{A spectral viewpoint: low-degree Fourier mass vs.\ parity}
\label{sec:spectral-view}

A useful complementary lens is Fourier analysis on the Boolean hypercube.
Let $f:\bits^{n}\to\mathbb{R}$ and define the Walsh--Hadamard characters
\[
\chi_S(x) \;=\; (-1)^{\sum_{i\in S} x_i}, \qquad S\subseteq[n].
\]
The set $\{\chi_S\}_{S\subseteq[n]}$ is an orthonormal basis with respect to the uniform distribution on $\bits^n$.
The Fourier coefficients are
\[
\hat f(S)\;=\;\mathbb{E}_{x\sim U(\bits^n)}\bigl[f(x)\chi_S(x)\bigr],
\qquad\text{and}\qquad
f(x)=\sum_{S\subseteq[n]}\hat f(S)\chi_S(x).
\]
We refer to $|S|$ as the \emph{Fourier degree} (or ``frequency'') of the coefficient $\hat f(S)$.

\begin{definition}[Fourier tail / spectral bandwidth]
\label{def:fourier-tail}
For $k\in\N$, define the degree-$>k$ Fourier tail
\[
\mathrm{Tail}_{>k}(f)\;=\;\sum_{S\subseteq[n]:\,|S|>k}\hat f(S)^2.
\]
For $\varepsilon\ge0$, we say $f$ is $(k,\varepsilon)$-\emph{spectrally band-limited} if $\mathrm{Tail}_{>k}(f)\le \varepsilon$.
\end{definition}

\paragraph{Parity is a pure high-frequency character.}
For any nonempty $S\subseteq[n]$, the parity function $\mathrm{PARITY}_S(x)=\chi_S(x)$ has Fourier expansion supported on a single set:
$\widehat{\mathrm{PARITY}_S}(S)=1$ and $\widehat{\mathrm{PARITY}_S}(T)=0$ for $T\neq S$.
Thus its spectral ``frequency'' is exactly $|S|$. More generally, if $f$ is $(k,\varepsilon)$-spectrally band-limited and $|S|>k$, then
\begin{equation}
\bigl|\mathbb{E}[f(x)\chi_S(x)]\bigr|^2=\hat f(S)^2\le\mathrm{Tail}_{>k}(f)\le\varepsilon.
\label{eq:tail-correlation}
\end{equation}
This elementary inequality is the link between concentration at low degree and small correlation with a high-degree parity.

\paragraph{$\AC^0$ has little high-frequency mass.}
A seminal theorem of Linial--Mansour--Nisan shows that $\AC^0$ functions have almost all Fourier mass on low degrees:
for $f$ computed by an $\AC^0_d$ circuit of size $M$, the tail above degree $k$ decays rapidly with $k$ (see \citep{linial1993constant}, and sharper bounds in \citep{tal2015tight}).
For fixed depth and polynomial size, correlation with full parity under the uniform distribution is negligible in $n$. Applying this statement to other subsets requires specifying their size relative to circuit depth and size, and it does not automatically apply under a nonuniform training distribution.

\paragraph{Local vs.\ global parity in NIM.}
In our encoding, the $j$th bit of the nim-sum is the parity of the $j$th bits across all $N$ heaps, hence a degree-$N$ character.
Conditional on a specified changed heap $i$, the sign encoding of the $j$th delta bit is the two-coordinate character $(-1)^{x_{i,j}+x'_{i,j}}$. When the changed index is unknown, the delta bit is specified on pairs of boards differing in at most one heap. A shallow extension to all pairs is
\begin{equation}
d_j(P,P')=\bigvee_{i=1}^{N}(x_{i,j}\oplus x'_{i,j}).
\label{eq:delta-extension}
\end{equation}
On the promised domain this is the required delta bit. For $N\ge2$, its sign encoding $(-1)^{d_j}$ is not a single character, even up to sign; for $N=1$ it is exactly a degree-two character. Thus local computability does not mean that the full unknown-index map has degree two. It follows from the explicit $\AC^0$ construction in Lemma~\ref{lem:nimber-diff-ac0}.

\paragraph{Connection to optimisation (informal).}
Fourier degree also aligns with a common empirical observation in deep learning: gradient-based training tends to fit low-frequency components before high-frequency ones (``spectral bias'' / ``frequency principle'') \citep{rahaman2019spectral,xu2019frequency}.
This suggests a possible analogy for parity learning, but continuous-domain spectral-bias observations do not prove how SGD learns Boolean parity or restoration under our nonuniform data distribution. We treat that connection as a hypothesis requiring separate measurements.

\subsection{From global invariants to local checks}
The core representational point is that the global invariant $\operatorname{nim}(P)$ is hard to compute from a single frame, whereas the transition quantity $\Delta(P,P')$ is locally computable from two frames (Lemma~\ref{lem:nimber-diff-ac0}).
Section~\ref{sec:supervised-restoration} isolates this effect by training one-frame and two-frame models on the same restoration targets.

\section{Experiments}
\label{sec:experiments}

The stronger experimental goal would be to show that an AlphaZero-style learner discovers effective play through self-play once a bypass is available. We do not establish that here. The theoretical barrier concerns what the restricted network can represent, independently of training; removing it leaves open whether the required computation can be learned, with what data and under which training regime.

We take a controlled intermediate step: supervised restoration learning from teacher--random transitions (Section~\ref{sec:supervised-restoration}) and a complementary single-frame multi-head shootout evaluation with imitated policy heads (Section~\ref{sec:exp-fsm-shootout}). These experiments supply guidance that outcome-only self-play would have to generate for itself. They test whether the proposed information and computation can be useful in training, while also exposing difficulties that remain after the input has been enriched.
Unless otherwise stated, both one-frame (1F) and two-frame (2F) agents share the same action parameterisation and network architecture (Section~\ref{sec:arch}).

\subsection{Neural architecture and training objective}
\label{sec:arch}

\subsubsection{State and action representation}
\label{sec:state-action-repr}
We use a fixed-width binary encoding of heap sizes with $k$ bits per heap. The single-frame input is the encoding of the current heap vector. The two-frame input concatenates the encodings of the previous and current heap vectors. When a previous frame is unavailable (e.g. on the first move of a game), we set the previous frame equal to the current position so that the input dimension is constant and the 2F policy remains well-defined.

The action space is parameterised as ``set heap $i$ to value $v$'' for $i\in\{1,\dots,N\}$ and $v\in\{0,1,\dots,2^k-1\}$, giving $N\cdot 2^k$ logits. The saved restoration evaluation takes an unmasked argmax over all logits; illegal predictions count as incorrect. Legality masking is used in the separate game-playing and FSM rollout routines. These are distinct evaluation conventions.

\subsubsection{Network architecture}
Both restoration agents use a policy-only MLP with two hidden layers and $N\cdot2^k$ output logits. There is no value head in the saved implementation. The two-frame input has dimension $2Nk$; the single-frame input has dimension $Nk$. The template and hidden width are shared, but parameter counts differ: the two-frame first layer adds $NkH=20{,}480$ parameters in the reported setting. All layers use ordinary floating-point arithmetic; these networks are not literal implementations of the FSQI gate model.

\begin{table}[h!]
\centering
{\footnotesize
\begin{tabularx}{\linewidth}{@{}lYYY@{}}
\hline
\textbf{Component} & \textbf{Type} & \textbf{Dimensions} & \textbf{Activation/}\allowbreak\textbf{Normalisation} \\
\hline
Input & State representation & Two-frame: $2Nk$ bits (1F: $Nk$) & None \\
Hidden 1 & Fully Connected & 1F: $Nk\to H$; 2F: $2Nk\to H$ & LayerNorm + ReLU \\
Hidden 2 & Fully Connected & $H\to H$ & ReLU \\
Policy Head & Linear Output & $H\to N\cdot2^k$ & Raw logits; Log-Softmax in loss \\
\hline
\end{tabularx}}
\caption{Neural network architecture used in the experiments. $H$ denotes the hidden width (a fixed hyperparameter).}
\label{table:arch}
\end{table}

\subsubsection{Training targets and loss}
For each training input (single-frame state or two-frame state-pair), the target is a soft distribution $\pi_{\mathrm{opt}}(\cdot\mid p')$ over optimal restoration moves.
The policy head is trained with cross-entropy to this target.
The soft-target cross-entropy also uses unmasked logits. No value loss is used in this supervised restoration experiment.

\subsubsection{Evaluation criteria in this version}
In this version we report two task-level notions of performance rather than full game-level Elo tournaments.
\emph{Move-level restoration performance} is measured by \textbf{functional correctness}: on a test position with nonzero nim-sum, a move is counted as correct if it produces a successor state with nim-sum zero.
Positions with zero nim-sum have no winning move (Theorem~\ref{thm:nim-win}); when reporting functional correctness we therefore condition on nonzero nim-sum positions.
\emph{Position-level evaluation performance} is measured by $P/N$ classification accuracy and $N$-position recall in FSM-controlled shootouts (Section~\ref{sec:exp-fsm-shootout}).

\subsection{Supervised restoration learning from teacher--random transitions}
\label{sec:supervised-restoration}
\label{sec:exp-restoration}

This experiment isolates the \emph{restoration response} as a supervised learning problem.
In normal-play impartial games, a restoration response is the move that returns play to a losing ($P$-)position after the opponent departs from it.
In Nim, this is equivalent to selecting an optimal (winning) move from the post-opponent position: if the previous position had nim-sum $0$, then after any opponent move the position is winning, and any optimal reply restores nim-sum $0$.

Motivated by Sections~\ref{sec:impossibility}--\ref{sec:multiframe}, the experiment asks a separate empirical question under a fixed supervised training protocol:
\emph{does access to the transition $(p,p')$ (two frames) enable successful restoration learning under this protocol, even when the one-frame model fails?}
This benchmark builds on the observation that parity-type invariants pose a challenge for reinforcement learning in impartial games \citep{zhou2022impartial,zhou2023parity}, while explicitly providing the transition information needed by Proposition~\ref{prop:restore}.

\subsubsection{Game and representations}
We consider Nim with $N=20$ heaps.
Each heap size is represented with $k=4$ bits (little-endian), so heap values lie in $\{0,\ldots,15\}$.
Moves are represented in a fixed action space of size $N\cdot 2^k = 20\cdot 16 = 320$:
an action chooses a heap index $i$ and a new heap value $v$. Legality requires $0\le v<h_i$; the reported unmasked metric counts violations as errors.

A \textbf{1-frame (1F)} policy receives only the current position $p'$ (80-bit input),
while a \textbf{2-frame (2F)} policy receives the concatenation $(p,p')$ of the previous and current positions (160-bit input).
Both agents use the shared MLP architecture described in Section~\ref{sec:arch} (two hidden layers of width $H=256$).

\subsubsection{Dataset generation (teacher--random--teacher)}
We generate a dataset of size $D=10^6$ examples of the form
\[
(p, p') \mapsto \pi_{\mathrm{opt}}(\cdot \mid p') ,
\]
where $\pi_{\mathrm{opt}}(\cdot \mid p')$ is a soft target distribution over optimal moves from $p'$.

Each example is generated as follows:
\begin{enumerate}
  \item Sample a random \emph{winning} Nim position $s$ (nim-sum $\ne 0$).
  \item Apply an optimal teacher move from $s$ to obtain $p$.
        For optimal play in Nim, $p$ has nim-sum $0$ (a $P$-position for the next player).
  \item Apply a uniformly random legal move from $p$ to obtain $p'$.
  \item Define the label $\pi_{\mathrm{opt}}(\cdot \mid p')$ as the uniform distribution over all optimal (winning) moves from $p'$.
\end{enumerate}
This construction ensures that each training input $(p,p')$ corresponds to the canonical restoration setting:
$p$ is a $P$-position, the random move produces a winning position $p'$, and the label is the set of restoration moves back to a $P$-position.

\paragraph{Why teacher--random--teacher?}
We do not sample arbitrary triples $(P,P',a)$ uniformly at random, because the restoration claim is about a specific interaction regime: the opponent departs a $P$-position and the agent must restore it. The teacher step ensures $P$ is a genuine $P$-position reachable under optimal play, the random move injects diversity in the departure $P\to P'$, and the final teacher move provides a correct restoration label for $P'$.

Compared to earlier empirical studies of impartial games \citep{zhou2022impartial,zhou2023parity}, we ``help'' the learner in two ways:
(i) we represent heap sizes in binary rather than as raw integers, and
(ii) we train on \emph{perfectly labelled} restoration targets (the full set of optimal moves), rather than requiring the agent to infer these targets from sparse win/loss outcomes.

We split the dataset into 80\% train / 10\% validation / 10\% test using a seeded permutation (seed 0), shared by the two representations. The teacher uses the first winning heap; this induces a particular, nonuniform distribution of predecessor positions. The results do not establish invariance to heap permutations, unseen board sizes, or arbitrary self-play histories.

\subsubsection{Training and checkpoints}
We train each model with supervised learning using cross-entropy to the soft target distribution.
We use Adam with learning rate $10^{-3}$ and batch size 512 for 200 epochs.
To study performance as a function of training, we save $C=10$ checkpoints per model at epochs
\[\{1,23,45,67,89,112,134,156,178,200\}.\]

\subsubsection{Evaluation metrics}
We evaluate restoration performance on held-out data:

\paragraph{Restoration accuracy (validation/test).}
Given a held-out example, the saved evaluation takes the argmax over all action logits and counts it as correct if it lies in the support of $\pi_{\mathrm{opt}}(\cdot\mid p')$. Illegal predictions therefore count as incorrect, and every legal optimal move is accepted.
The saved implementation averages batch accuracies equally, including the shorter final batch, rather than weighting them by batch size. The reported decimals are those saved batch means, not exact correct-example counts divided by 100,000. The same convention applies to the displayed validation curve.

\subsubsection{Results}

\begin{figure}[t]
\centering
\includegraphics[width=0.95\linewidth]{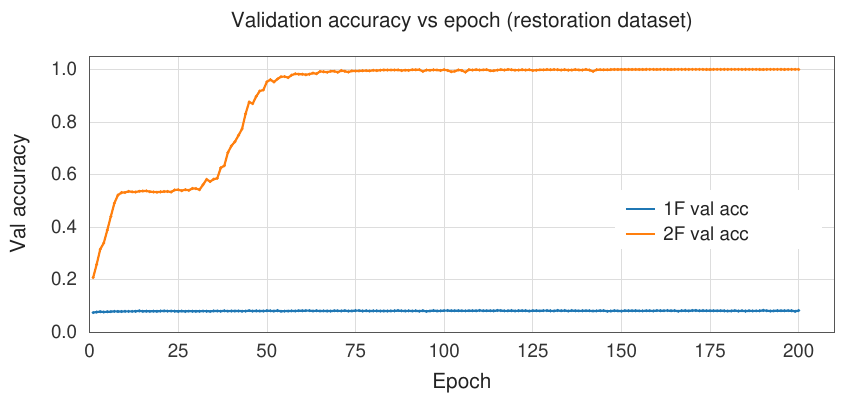}
\caption{Validation restoration accuracy vs.\ epoch on the restoration dataset ($N=20$, $k=4$, $D=10^6$).
In this saved run, the 1F model has low accuracy while the 2F model passes through an intermediate plateau around 0.54 before improving sharply towards near-perfect accuracy. The transition suggests a learning bottleneck that training eventually overcomes; it is distinct from the representational lower bound. Values use the saved batch-mean metric.}
\label{fig:valacc-H20k4}
\end{figure}

\begin{figure}[t]
\centering
\includegraphics[width=0.95\linewidth]{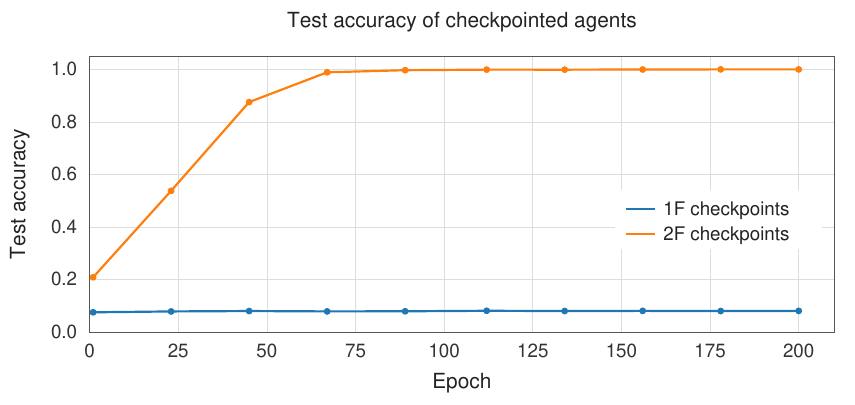}
\caption{Test restoration accuracy of checkpointed agents ($N=20$, $k=4$, $D=10^6$).
Checkpoints are saved at epochs $\{1,23,45,67,89,112,134,156,178,200\}$. The 2F checkpoints confirm the improvement beyond the intermediate plateau; the more densely sampled validation curve in Figure~\ref{fig:valacc-H20k4} shows its timing.}
\label{fig:testacc-H20k4}
\end{figure}

\paragraph{One-frame does not learn restoration; two-frame eventually does.}
Figure~\ref{fig:valacc-H20k4} shows validation restoration accuracy as a function of epoch.
The 1F model has low test restoration accuracy throughout training, approximately 0.075--0.081 across saved checkpoints. We do not identify this range with a measured random-legal-move baseline, which is not supplied in the reported evaluation.
In contrast, the 2F model learns the restoration mapping and reaches essentially perfect accuracy with enough training.
On the test set (Figure~\ref{fig:testacc-H20k4}), the 2F checkpoints achieve 0.209 at epoch 1, 0.538 at epoch 23, 0.876 at epoch 45, 0.989 at epoch 67, and 0.99998 by epoch 178 (remaining $\ge 0.99993$ at epoch 200).

\paragraph{Learning in stages: the plateau and subsequent improvement.}
The plateau near $0.54$ and later rise in Figures~\ref{fig:valacc-H20k4}--\ref{fig:testacc-H20k4} suggest a temporary learning bottleneck even with two-frame input. The learner makes substantial progress, remains at an intermediate level, and then improves sharply. This is evidence about training dynamics within the saved run, rather than a further circuit lower bound.

A diagnostic of the saved checkpoints makes the remaining difficulty more concrete. For a predicted move setting heap $i$ to $v$, compare its proposed change $p'_i\oplus v$ with the required nim-sum of $p'$. On the reconstructed and independently verified 100,000-position test set, the epoch-23 model makes only legal moves and gets the upper three bits of this change jointly correct on 99.998\% of examples. Of its 46,206 failed moves, 46,204 differ from the correct change only in the lowest bit. At epoch 45 all 12,437 errors are confined to that bit; at epoch 178 only two errors remain, both of the same kind. These diagnostic counts weight examples equally, unlike the saved batch-mean accuracies plotted above.

Thus the later improvement is accounted for at the output level by learning to select moves with the correct lowest bit while retaining the already accurate upper bits. This supports staged acquisition of the restoration behaviour. It does not identify an internal representation or explain why optimisation follows this order. Nor does the plateau's proximity to $8/15$ establish a random-bit guessing mechanism: that baseline assumes a particular distribution and prediction rule. Internal probes, interventions and independent training runs are needed to explain the transition and test its reproducibility.

\paragraph{Scope and provenance of the reported result.}
This controlled supervised run shows a large advantage for transition information at $N=20$, $k=4$, $D=10^6$ and seed 0. It does not prove worst-case mastery, efficient self-play learning, or the asymptotic circuit lower bound. The checkpoint CSV is identified by run suffix \texttt{20260214\_120801\_cd18fc9f}; the architecture and evaluation convention above follow its associated saved notebook.
We report this documented configuration and do not make an unsupported claim of robustness across a hyperparameter or board-size sweep. Multiple independent seeds, matched-parameter controls and out-of-distribution evaluations remain useful follow-up experiments.

\subsection{Single-frame evaluation via FSM-controlled shootouts (multi-head)}
\label{sec:exp-fsm-shootout}

\paragraph{Purpose.}
We complement the two-frame restoration study (Section~\ref{sec:exp-restoration}) with an experiment targeting the \emph{single-frame} bypass from Proposition~\ref{prop:rollout-sufficiency}:
learned approximations to the structured rollout policies combined with a finite-state shootout controller (Figure~\ref{fig:fsm-shootout-controller}). The trained floating-point networks are not claimed to satisfy the formal $\AC^0$ gate restrictions.
The goal is to test the \emph{rollout interface} itself (how much information can be extracted by controlled shootouts), not to claim that self-play discovers the structured heads without additional inductive bias.

\paragraph{Setup.}
We consider NIM with $N=20$ heaps and bit-width $k=4$ (so $B=k=4$ heads), matching the encoding of Section~\ref{sec:state-action-repr}.
We evaluate on a balanced test set of 2000~positions with $50\%$ $P$-positions ($\sigma=0$) and $50\%$ $N$-positions ($\sigma\neq 0$).

\paragraph{Training the rollout heads.}
We train a shared-trunk network with $B=4$ policy heads by imitation learning against the structured policies $\pi_0,\dots,\pi_{B-1}$ from Proposition~\ref{prop:rollout-sufficiency}.
Each head~$\pi_j$ subtracts~$2^j$ from the leftmost heap of size at least~$2^j$ (falling back to subtracting~1 from the leftmost nonzero heap).
When multiple heaps qualify for the same phase of~$\pi_j$ (e.g.\ several heaps have $h_i \geq 2^j$), any choice among them produces a rollout with the same total move count and hence the same parity outcome.
We therefore train with \emph{soft targets}: for each position and head~$j$, the label is uniform over all equivalent actions, and the loss is cross-entropy against this soft distribution.
This avoids penalising the model for an arbitrary tie-breaking choice among equally valid heaps.

The network uses the same shared-trunk architecture as Table~\ref{table:arch} (two hidden layers of width $H=256$ with LayerNorm), with $B$ independent linear output heads each producing $N \cdot 2^k = 320$ action logits.
Training uses Adam (learning rate $10^{-3}$, batch size 1024) for 12 epochs on $2\times10^5$ imitation examples, with a separate 20,000-example validation set and seed 7 in the saved notebook.

\paragraph{Rollout evaluation.}
Each rollout is \emph{deterministic}: both players follow the same greedy policy head~$\pi_j$ (argmax over legal actions) until termination.
The rollout outcome is~1 if the first player wins (makes the last move) and~0 otherwise.
Since rollouts are deterministic, a single rollout per (position, head) pair suffices.

The per-head accuracy metric is whether the model's rollout outcome matches the teacher's rollout outcome---the correct channel bit---not whether it matches the $P/N$ label directly.
Each \emph{exact teacher} head computes a specified linear functional ($r_0=\sigma_0$; $r_j=\sigma_j\oplus\sigma_0$ for $j\ge1$). The learned heads can make channel errors. The FSM uses an OR of the queried outcomes for $P/N$ classification, which is a different task from reconstructing every nim-sum bit.

\paragraph{FSM-controlled shootout evaluation.}
Given a test position~$P$, the evaluator runs the controller of Figure~\ref{fig:fsm-shootout-controller}:
it queries head~$\pi_0$ first; if $r_0 = 1$ it immediately declares $\sigma(P) \neq 0$ (N-position).
Otherwise it queries heads $\pi_1, \pi_2, \ldots, \pi_{B-1}$ sequentially, declaring N-position as soon as any $r_j = 1$, or declaring P-position if all $B$~bits are~0.
Early stopping reduces the average number of rollouts below~$B$ on N-positions.
We report (i) overall $P/N$ evaluation accuracy and (ii) mean number of shootouts used.

\paragraph{Results.}
Table~\ref{tab:fsm-shootout} reports the results.

\emph{Multi-head FSM achieves perfect classification.}
The saved output reports correct $P/N$ classification on all 2,000 test positions, matching the teacher's final decisions. However, the saved per-head channel accuracies are $2000/2000$, $2000/2000$, $1469/2000$ and $1480/2000$, respectively: $1.0000$, $1.0000$, $0.7345$ and $0.7400$.
Consequently, this result does not establish exact channel recovery or recovery of the full invariant by the learned model. Early stopping leaves some channels unqueried, and even queried channel errors need not change the final OR decision. These are transcribed saved outputs, not a fresh checkpoint evaluation; reached-head error counts and a reproducible end-to-end rerun are needed for a stronger claim.

\emph{Single-head baseline.}
The single-head baseline achieves 0.768~overall accuracy and 0.536~N-position recall. These numbers have an exact theoretical explanation. With $k=4$ bits, an N-position has nim-sum $\sigma \in \{1,\dots,15\}$.
Of these 15~values, exactly~8 are odd ($\sigma_0 = 1$: the values $1,3,5,7,9,11,13,15$) and 7~are even ($\sigma_0 = 0$).
Head~$\pi_0$ detects an N-position only when $\sigma_0 = 1$, so the theoretical N-position recall is $8/15 \approx 0.533$ (observed: 0.536 on the finite test set). All P-positions are correctly classified (since $\sigma = 0$ implies $\sigma_0 = 0$), so on a balanced test set the overall accuracy is
$\tfrac{1}{2}\cdot 1 + \tfrac{1}{2}\cdot\tfrac{8}{15} = \tfrac{23}{30} \approx 0.767$
(observed: 0.768).
This calculation describes the particular classifier that declares an N-position exactly when $r_0=1$. It is not the winning-move success bound of Theorem~\ref{thm:search-gap}(i), and does not establish a limit on arbitrary classifiers or rollout policies.

\emph{Different metrics.}
The recall $8/15$ measures the fraction of nonzero nim-sums detected by the exact least-significant-bit channel. Restoration accuracy instead measures whether a complete selected move cancels the full nim-sum. A resemblance between the recall and a training plateau is not evidence that the same learned mechanism produces both.

\emph{Early stopping.}
The average number of shootouts is 2.87 rather than the worst-case~4, because any N-position with $\sigma_0 = 1$ is caught at head~0 (one rollout); only those with $\sigma_0 = 0$ require further heads.
\emph{Scope.}
We report the saved seed-7 configuration above. Perfect classification on this finite test set does not establish all-position correctness, robustness across seeds, or successful self-play discovery. We make no broader sweep claim without corresponding run records.

\begin{table}[t]
\centering
\small
\begin{tabular}{p{0.36\linewidth}cccc}
\hline
Method & Heads & Rollouts & Accuracy & Recall \\
\hline
Single head (model) & 1 & 1.00 & 0.768 & 0.536 \\
Multi-head + FSM (model) & 4 & 2.87 & 1.000 & 1.000 \\
Single head (teacher) & 1 & 1.00 & 0.768 & 0.536 \\
Multi-head + FSM (teacher) & 4 & 2.87 & 1.000 & 1.000 \\
\hline
\end{tabular}
\caption{Single-frame $P/N$ evaluation by FSM-controlled shootouts ($N=20$, $k=4$), reporting overall $P/N$ accuracy and $N$-position recall.}
\label{tab:fsm-shootout}
\end{table}


\paragraph{Reproducibility status.}
The restoration notebook, checkpoint-accuracy CSV and FSM notebook are retained with the local review materials. The public repository previously named for this work does not currently provide a reproducibility package; we therefore do not claim that code and data for all configurations are publicly available. The numerical results above remain tied to the saved runs and their stated limitations. A public versioned package and independently repeated evaluation are future work.

\section{Discussion and broader implications}
\label{sec:discussion}

The three bypasses identify different resources for handling a global invariant: multi-frame inputs expose local transition information, multiple rollout heads provide complementary channels, and persistent recurrence accumulates the invariant through sequential updates. In the recurrent route, memory connects successive positions while the growing processing horizon supplies the depth needed for parity. The common design principle is to match the information and computation available to the agent to the invariant it must maintain. We briefly discuss directions beyond the specific NIM setting.

\subsection{Theoretical idealisation vs.\ practical architecture}
\label{sec:idealisation}
The simulation theorem concerns Boolean literal-threshold gates with nonnegative weights and bounded thresholds, including their permitted compositions. The experiments use floating-point signed linear layers and ReLU as well as LayerNorm and a Log-Softmax loss. Their difference from the theorem therefore extends beyond normalisation layers.
A failure at one finite board size does not establish an asymptotic representational impossibility; success on a finite dataset does not establish strong mastery. The controlled comparison supports a narrower empirical conclusion: the reported transition representation is much easier to learn under this supervised protocol.

\subsection{Local information in combinatorial search}
\label{sec:math}
Exposing local invariants may also help guide combinatorial search. Our work on large Condorcet domains used structural constructions and local restrictions \citep{karpov2023setalternating,zhou2024cdl,zhou2024phd}. This is a motivating analogy, not an $\AC^0$ reduction or a demonstrated parity lower bound for that search problem. Transferring the present argument requires a precisely encoded computational task.

\subsection{Beyond NIM: harder impartial games as benchmarks}
\label{sec:othergames}

By Sprague--Grundy theory, every finite impartial game position has a nimber. However, the computational complexity of determining nimbers varies widely. Some impartial games have PSPACE-complete winner problems, such as directed vertex Geography \citep{schaefer1978games}. Unless $\mathrm P=\mathrm{PSPACE}$, no polynomial-time algorithm solves all such instances. Having a nimber does not supply an efficient way to compute it.

NIM is therefore a diagnostic starting point: its efficient optimal strategy lets us separate a representational obstruction from the difficulty of discovering a strategy. The longer-term benchmark is self-play on impartial games whose strategic structure is only partly understood or whose exact analysis becomes impractical as positions grow. Three candidates illustrate different challenges:

\begin{itemize}[leftmargin=*]
\item \textbf{Sprouts.} Players join spots by noncrossing curves, add a spot on each new curve, and keep every spot's degree at most three. The normal-play outcome conjecture for the standard $n$-spot opening remains open despite substantial progress with exact solvers \citep{cizek2026sprouts}. Learning to exploit the evolving region structure, and to choose winning moves in intermediate positions, would be a stronger test than predicting opening outcomes from $n$ alone.
\item \textbf{Chomp.} A move removes a chosen square and all squares above and to its right; taking the distinguished corner loses. Strategy stealing establishes a first-player win on every nontrivial rectangle, but does not supply the winning moves \citep{berlekamp2001winningways}. Larger boards and their irregular intermediate positions offer a challenge beyond solved special families; the analysis of Chomp continues to pose open problems \citep{nowakowski2025unsolved}. The benchmark should test move selection and generalisation across shapes, not just the known opening winner.
\item \textbf{Grundy's game.} A move splits one heap into two nonempty unequal heaps. Whether its nim-value sequence is eventually periodic remains an open problem \citep{nowakowski2025unsolved}. This offers a test of whether self-play discovers useful structure in component values and how they combine. Testing unseen heap sizes and sums of heaps can distinguish such progress from memorising a finite table.
\end{itemize}

These are proposed learning challenges, not consequences of the present $\AC^0$ lower bound. An unresolved structural question is not itself a complexity lower bound, and worst-case hardness need not make a chosen instance distribution difficult. A useful evaluation would reserve unseen sizes and positions, report both training and decision-time computation, and compare play with exact solvers where feasible and strong search baselines elsewhere. Beyond the range of exact verification, improved win rates are evidence of stronger play rather than a solution of an open mathematical problem. The central question is whether self-play discovers reusable strategic structure that extends the range of positions it can handle under a fixed computation budget.

\subsection{What the rollout separation establishes}
\label{sec:multipolicy-discussion}
Theorem~\ref{thm:search-gap}(i) bounds winning-move selection for fixed bit-width, a specified family of root/one-heap measurements and a polynomial-size, constant-depth readout. It does not bound general single-policy search. Within that interface, the $B$ structured channels reconstruct the invariant and permit direct move selection.
The extra computational resource is the sequential rollout, not the number of neural output coordinates alone. Repeated deterministic measurements of the same affine channel cannot create additional channel bits, but changing the permitted query transformations or controller computation changes the model and can remove that restriction.

\subsection{Learning bottlenecks after the representational barrier}
\label{sec:opt-gap}

The constructions establish that the required computations can be expressed using the stated resources: a shallow restoration rule, structured rollout policies, or a shallow recurrent update iterated over the board. This removes one obstacle before training begins. It does not determine whether gradient descent will find those computations, whether sparse game rewards provide enough guidance, or whether the resulting behaviour generalises.

The restoration experiment makes this distinction tangible. Even with transition information and correct supervision, learning passes through an intermediate plateau before reaching near-perfect measured accuracy. The checkpoint diagnostic in Section~\ref{sec:supervised-restoration} associates the later improvement with the lowest bit of the selected move's XOR change. We therefore have empirical evidence of a remaining learning bottleneck that is largely overcome in this run, alongside a theorem about a different, representational obstruction. The data do not establish how many distinct learning mechanisms produce the stages or why the transitions occur.

The immediate experimental question is whether AlphaZero-style self-play can discover the restoration rule, complementary rollout policies or recurrent accumulator without their supervised targets. A follow-up should compare outcome-only training with curricula and auxiliary supervision, and examine the learned computations as well as win rates. For recurrence, this includes accurate accumulation and reliable state retention across separators. These controlled NIM tests would establish a basis for the harder benchmarks in Section~\ref{sec:othergames}; efficient self-play learning and the training regimes needed to achieve it remain open.

\subsection{Significance and impact}
Our results suggest several broader implications.

\begin{itemize}[leftmargin=*]
\item \textbf{For learning theory:} representational sufficiency and learnability are separate questions. A promised predecessor state, complementary measurements, or sequential computation can each alter the resources available for an invariant-based decision.
\item \textbf{For neural architecture:} short history windows (or difference features) can be a principled way to bypass parity-like barriers without scaling depth/width. When history is unavailable, multiple rollout heads offer an alternative within a single-frame architecture. When growing processing time is available, persistent recurrence offers a third route, building the invariant sequentially and retaining it across positions.
\item \textbf{For practice:} circuit complexity can help formulate architectural hypotheses, provided the implemented model satisfies the mathematical assumptions. Tests of those hypotheses should distinguish representation, training dynamics, task distribution and computation budget.
\end{itemize}


\section{Conclusion}
\label{sec:conclusion}
A fixed-depth Boolean literal-threshold model with nonnegative fixed-grid weights and bounded thresholds is simulable in $\AC^0$. The NIM parity reduction then rules out strong single-frame mastery in that model.

Three constructions identify resources beyond that restricted inference problem. Two-frame access supports constant-depth restoration when the preceding position has nim-sum zero. Distinct structured rollout policies reveal complementary invariant bits; a single-channel limitation holds only under our explicit query and readout restrictions. Persistent recurrence implements running XOR over heap rows, exchanging fixed total latency for a growing sequence of simple updates. Cross-position memory additionally supports comparisons, while replay or board storage supplies the heaps needed for move selection.

The saved floating-point experiments show a large supervised two-frame restoration advantage and perfect finite-test $P/N$ classification by a learned multi-head controller. The latter coexists with imperfect individual channel recovery. The restoration plateaus and checkpoint diagnostics also show that learning can proceed in stages after transition information is supplied. These observations do not certify the formal gate assumptions, worst-case mastery, or self-play discovery.

The contribution is a separation of the information and computation needed for a decision from the training process that must discover it. Establishing when self-play learns the restoration rule, complementary rollout policies or recurrent accumulator remains a distinct experimental question.

\section{Development, verification and author responsibility}
\label{sec:development}
\label{sec:verification}
The earlier version of this paper was developed solely by the author, with editorial feedback from AI assistants. The present version was developed interactively with the AI assistants Astra and Claude. Extensive feedback from the Lean formalisation and its audits helped to improve the presentation, clarify definitions and hypotheses, and make the scope of the claims more precise. The author takes sole responsibility for the paper, including its arguments, interpretation and use of AI assistance.

The accompanying Lean development \citep{riis2026lean} connects the mathematical results to explicit Boolean circuits, fixed-scale networks and complete game strategies. It checks the parity reduction, the finite probability bound and its attainability, move selection with binary outputs, the rollout executors and both shootout schedules, and the recurrent protocols with replay or board storage. The general weak-mastery definition, Fourier expansion and the real-parameter noise calculation are also formalised. An independent rebuild checks 1,588 theorem and lemma declarations across 38 local Lean files, comprising 24,493 source lines; these counts include auxiliary results and exclude Lean and Mathlib themselves.

The boundary of this verification is explicit. Classical complexity results, notably H{\aa}stad's parity hardness and correlation bounds and the Linial--Mansour--Nisan/Tal spectral bounds, remain external inputs \citep{hastad1986thesis,hastad2014correlation,linial1993constant,tal2015tight}. The single-channel result is checked conditional on a stated negligible parity-correlation theorem, including its application to the readout and its probability calculation. The spectral discussion uses a separate Fourier-tail premise; its connection to the exact conventions of a published tail theorem is not part of the checked development. The conditional implication from a tail bound to small parity correlation is checked. Empirical measurements and training runs are supported by experimental code and saved results, rather than by Lean proofs.

The author's review of the formalisation focused on the interface between the paper and Lean: whether the formal definitions, hypotheses and conclusions express the intended mathematical claims. The author did not manually inspect every theorem or lemma declaration or every line of Lean source. Mechanical checking verifies the formal proofs relative to their definitions and stated hypotheses; it does not independently certify the correspondence with the prose, the suitability of the modelling assumptions, or the experimental results.

The accompanying interactive edition \citep{riis2026interactive} reproduces the paper text and links individual statements to their definitions, hypotheses and proofs, with optional explanations alongside them. The paper's arguments remain self-contained at the level of its stated classical inputs; the interactive material supplies further detail rather than replacing a proof needed to understand a claim. Stable statement labels and a versioned correspondence record identify the formal results and any external assumptions. The edition includes the complete local Lean sources, pinned dependency information and reproducible build instructions. Claude generated the interactive edition and its link structure, which underwent a detailed, extensive audit by Astra. The author extensively cross-checked the edition but did not inspect every internal link and asset reference individually. An automated audit of this version verified the targets of 50,970 such references, including repeated occurrences. These navigation checks do not by themselves establish that each link provides the intended explanation or proof.

Neither the mathematical verification nor the experiments establish that gradient descent or self-play will discover the constructions in general.


\noindent\begin{minipage}{\linewidth}
\acks{The author received no dedicated funding for this work.}
\end{minipage}

\appendix

\section{Formal definitions of fixed-scale quantised neural models}
\label{app:defs}

This appendix provides formal definitions of the fixed-scale quantised neural models used throughout the paper.

\subsection{Preliminaries: fixed-scale quantisation and representable parameters}
\label{app:precision}

Fix constants $W,D\in\N$.
Define the fixed finite set of representable parameters
\[
\Qset_{W,D}=\Big\{\frac{\ell}{D}: \ell\in\{0,1,\dots,W\}\Big\}.
\]
In particular, \emph{the set of allowable weights/thresholds does not depend on input length $n$}.
This models \emph{fixed-scale quantisation with bounded range}: values lie on a fixed grid (step $1/D$) and saturate at magnitude $W/D$, independent of the input length $n$.
It is a restricted Boolean gate abstraction; neither floating-point nor general signed fixed-point networks automatically satisfy it.
In particular, mere finiteness of a floating-point format is not what drives our $\AC^0$ containment; the key is that the effective range/scale of thresholds does \emph{not} grow with fan-in as $n$ increases.

We will allow negations on wires (equivalently, $\AC^0$ has NOT gates). This is standard in circuit complexity and avoids conflating two independent issues: (i) parity hardness vs (ii) monotonicity restrictions.

\subsection{Feedforward model}
\label{app:NN}

\begin{definition}[Literal-threshold network]
\label{def:NN-formal}
A fixed-scale quantised $\NN$ of depth $L$ on $n$ Boolean inputs is specified by:
\begin{itemize}[leftmargin=*]
\item Layer widths $m_0=n$, $m_1,\dots,m_L$, where $m_\ell\le \poly(n)$.
\item For each layer $\ell\in\{1,\dots,L\}$ and unit $i\in\{1,\dots,m_\ell\}$:
\begin{itemize}[leftmargin=2em]
\item weights $w_{\ell,i,j}\in \Qset_{W,D}$ for $j\in\{1,\dots,m_{\ell-1}\}$,
\item a threshold $\theta_{\ell,i}\in \Qset_{W,D}$,
\item an optional negation bit $\eta_{\ell,i,j}\in\{0,1\}$ indicating whether input $j$ is negated before weighting.
\end{itemize}
\end{itemize}
The computation proceeds as follows. Let $y^{(0)}=x\in\bits^n$. For $\ell\ge 1$,
\[
y^{(\ell)}_i \;=\;
\Heav\!\Big(\sum_{j=1}^{m_{\ell-1}} w_{\ell,i,j}\cdot (y^{(\ell-1)}_j \oplus \eta_{\ell,i,j}) \;-\; \theta_{\ell,i}\Big),
\]
where $\Heav(z)=1$ if $z\ge 0$ and $\Heav(z)=0$ otherwise.
The network output is $y^{(L)}$ (or a designated subset of its coordinates).
\end{definition}

\subsection{Recurrent model (finite window)}
\label{app:RNN}

\begin{definition}[Bounded-unrolling recurrence]
\label{def:RNN-formal}
Fix a time window $T\in\N$.
A fixed-scale quantised $\RNN$ consists of:
\begin{itemize}[leftmargin=*]
\item input bits $x_t\in\bits^n$ at time $t$,
\item a hidden state $h_t\in\bits^{m}$ for some $m\le \poly(n)$,
\item an output $o_t\in\bits^{k}$ for $k\le \poly(n)$,
\item fixed-scale quantised parameters (weights/thresholds in $\Qset_{W,D}$ and optional input negations) shared across time,
\end{itemize}
such that each update $h_{t+1}$ and $o_t$ is computed by a depth-$L$ fixed-scale quantised threshold network over the current input and earlier states within the window.
Each inference starts from an input-independent initial state, and the entire computation producing its output contains at most $T$ recurrent updates. No boundary state may carry additional input-dependent computation from before that window.
Unrolling this bounded computation yields a feedforward computation of depth $O(L\cdot T)$.
A bound only on the lag between connected states does not imply this bound on total unrolling; the persistent model in Section~\ref{sec:recurrent-bypass} deliberately allows more than $T$ updates.
\end{definition}

\subsection{Attention-style model (finite window)}
\label{app:LTST}

\begin{definition}[Gate-realised attention-style model]
\label{def:LTST-formal}
Fix a time window $T\in\N$.
A fixed-scale quantised $\LTST$ model processes a sequence $(x_{t-T+1},\dots,x_t)$ by:
\begin{itemize}[leftmargin=*]
\item computing short-term and long-term attention scores using fixed-scale quantised parameters in $\Qset_{W,D}$,
\item combining attended summaries through a constant number of layers of fixed-scale quantised threshold computations,
\item producing outputs $o_t\in\bits^k$.
\end{itemize}
Every score computation, input-dependent coefficient selection, encoding and value aggregation is required to have a polynomial-size, constant-depth implementation in the Boolean literal-threshold gates of Definition~\ref{def:NN-formal}. Intermediate state is Boolean, and total unrolling is bounded as in Definition~\ref{def:RNN-formal}. Under these explicit implementation assumptions, the simulation theorem applies by composition. Merely requiring parameters or coefficients to lie in a finite set is insufficient. This extension makes no claim about standard softmax attention without such a reduction.
\end{definition}

\begin{remark}[Scope of the simulation]
Theorem~\ref{thm:sim-ac0} applies because $L,T,W,D$ are constants and each unit’s parameters come from a finite set independent of $n$.
Appendix~\ref{app:ac0proof} gives an explicit circuit construction.
\end{remark}

\section{An \texorpdfstring{$\AC^0$}{AC0} construction for simulating fixed-scale quantised networks}
\label{app:ac0proof}

This appendix gives a clean circuit construction proving Theorem~\ref{thm:sim-ac0}.

\subsection{One threshold unit is in \texorpdfstring{$\AC^0$}{AC0} under fixed-scale quantisation}
\label{app:oneunit}

We first show that a single fixed-scale quantised threshold unit (as in Definition~\ref{def:NN-formal}) is computed by a depth-2 $\AC^0$ circuit of polynomial size.

\begin{lemma}[Fixed-scale quantised threshold gate $\in\AC^0$]
\label{lem:unit-ac0}
Fix constants $W,D\in\N$.
Let $z_1,\dots,z_m\in\bits$ be Boolean inputs, and let $w_1,\dots,w_m,\theta\in\Qset_{W,D}$.
Consider the Boolean function
\[
f(z_1,\dots,z_m) \;=\; \Heav\!\Big(\sum_{j=1}^m w_j z_j - \theta\Big).
\]
Then $f$ is computable by a depth-2 $\AC^0$ circuit of size $O(m^K)$, where $K:=W$ is a constant (depending only on $W,D$).
\end{lemma}

\begin{proof}
Multiply the inequality by $D$ to clear denominators.
Write $w_j=a_j/D$ and $\theta=b/D$ with $a_j\in\{0,1,\dots,W\}$ and $b\in\{0,1,\dots,W\}$.
Then
\[
f(z)=1 \quad\Longleftrightarrow\quad \sum_{j=1}^m a_j z_j \;\ge\; b.
\]
If $b=0$ the function is identically $1$. Assume $b\ge 1$.

Because each $a_j$ is a nonnegative integer and each active input contributes at least $1$ to the left-hand side whenever $a_j\ge 1$, any satisfying assignment contains a subset of at most $b$ active variables whose coefficients already sum to at least $b$.
More formally: if $\sum_j a_j z_j \ge b$, let $S=\{j:z_j=1 \text{ and } a_j\ge 1\}$. Choose any multiset-free subset $T\subseteq S$ by greedily adding indices until the partial sum reaches $\ge b$. Since each added index increases the sum by at least $1$, this process stops after at most $b$ steps. Because $b\le W$, we get $|T|\le W=:K$.

Therefore,
\[
f(z)=1 \quad\Longleftrightarrow\quad
\bigvee_{\substack{T\subseteq[m]\\ |T|\le K\\ \sum_{j\in T} a_j\ge b}}
\ \bigwedge_{j\in T} z_j.
\]
This is a depth-2 DNF: an OR of ANDs. Here depth counts AND/OR layers; counting input negations as gates adds at most one layer, as in the implementation bounds below.
The number of terms is at most $\sum_{r=0}^K \binom{m}{r}=O(m^K)$, polynomial since $K$ is constant.
Unbounded fan-in AND/OR gates give an $\AC^0$ circuit of depth $2$ and size $O(m^K)$.
\end{proof}

\begin{remark}[Negations]
If some inputs are negated before weighting, i.e.\ $(z_j\oplus \eta_j)$, this simply replaces $z_j$ by either $z_j$ or $\neg z_j$ in each conjunction term.
Since $\AC^0$ allows NOT gates, Lemma~\ref{lem:unit-ac0} still applies.
\end{remark}

\subsection{From one unit to a full network}
\label{app:network}

\begin{lemma}[One layer of fixed-scale quantised units is in $\AC^0$]
\label{lem:layer-ac0}
Fix $W,D$ and let a layer consist of $m'$ threshold units whose inputs are outputs of a previous layer of width $m$, with $m,m'\le \poly(n)$.
Then the layer can be simulated by an $\AC^0$ circuit of constant depth and polynomial size.
\end{lemma}

\begin{proof}
Apply Lemma~\ref{lem:unit-ac0} independently to each unit. Each unit yields a depth-2 circuit of size $O(m^K)$.
Combine these $m'$ circuits in parallel; the total size is $m'\cdot O(m^K)=\poly(n)$ and depth remains $2$.
\end{proof}

\begin{proof}[Proof of Theorem~\ref{thm:sim-ac0}]
For a feedforward $\NN$ of constant depth $L$, repeatedly replace each layer by the $\AC^0$ circuit from Lemma~\ref{lem:layer-ac0}, wiring outputs of layer $\ell-1$ circuits as inputs to layer $\ell$ circuits.
Depth multiplies by a constant factor: overall depth is $O(L)$ (constant), and size remains polynomial as a composition of polynomial-size layers.

For $\RNN$ and $\LTST$ with time window $T$, unroll the computation for $T$ steps (constant) to obtain a feedforward computation of depth $O(LT)$ (still constant). Apply the same layer-by-layer replacement.
\end{proof}

\section{Implementation details: shallow circuits and fixed-scale networks}
\label{app:impl}

The constructions share a small collection of Boolean operations. Each heap is a $B$-bit block, and a move is a validity bit, a binary heap index, and a $B$-bit replacement value. The circuits below have polynomial size in their supplied input length. Their depth does not grow with the number of heaps or bits. All XORs described as shallow here combine \emph{two} bits, or a fixed number under a bounded-change promise; an XOR of arbitrarily many heap rows is never treated as a primitive gate.

\paragraph{From a game rule to a network.}
\label{par:construction-example}
The networks in this appendix are constructed explicitly from the game rules: their connections and weights are specified by an algorithm, rather than found by training. To see what the construction computes, revisit Figure~\ref{fig:restoration-example}. The preceding board $Q=(3,5,6)$ has nim-sum zero; after the opponent changes $5$ to $4$, the current board is $Q'=(3,4,6)$. Comparing the frames gives $\delta=5\oplus4=1$. In parallel, the circuit proposes replacements $(3\oplus1,4\oplus1,6\oplus1)=(2,5,7)$ and tests which decrease their heaps. Only $2<3$ passes, so the output selects heap index $0$ with replacement value $2$, giving $(2,4,6)$ with nim-sum zero. For $B=3$ and a two-bit index, the output fields are validity $1$, index $00$, and value $010$.

\begin{figure}[htbp]
\centering
\begin{tikzpicture}[font=\small, box/.style={draw,rounded corners,align=center,text width=4.1cm,minimum height=1.05cm}, >=Stealth]
\node[box] (a) {Two frames\\$Q=(3,5,6)$, $Q'=(3,4,6)$};
\node[box,right=0.5cm of a] (b) {Extract the difference\\$\delta=1$};
\node[box,right=0.5cm of b] (c) {Propose replacements\\$(2,5,7)$};
\node[box,below=0.6cm of c] (d) {Test decreases in parallel\\$(\mathrm{true},\mathrm{false},\mathrm{false})$};
\node[box,left=0.5cm of d] (e) {Select the first candidate\\heap $0$, value $2$};
\node[box,left=0.5cm of e] (f) {Encode the move\\$1\mid00\mid010$};
\draw[->] (a) -- (b); \draw[->] (b) -- (c); \draw[->] (c) -- (d);
\draw[->] (d) -- (e); \draw[->] (e) -- (f);
\end{tikzpicture}
\caption{A schematic of the checked two-frame construction on one legal transition. Boxes group operations, not individual network layers. The general selector also handles multiple candidates, fallback moves and terminal boards.}
\label{fig:construction-pipeline}
\end{figure}

The same move-selection operations are reused when rollout outcomes supply the target nim-sum. In the recurrent construction the target is accumulated over rows, and replay supplies the rows again for candidate testing. These interfaces explain how the constructions avoid asking one shallow network to extract parity from an arbitrary single board. More heaps require more parallel circuitry; executing a rollout or a recurrent scan still requires repeated evaluations.

\paragraph{Local differences and candidate moves.}
On a legal transition only one heap changes. Consequently, bit $k$ of the delta is
\[
\delta_k=\bigvee_{i=0}^{N-1}(x_{i,k}\oplus x'_{i,k}).
\]
The single-change promise makes OR equal to the required XOR across changed rows. For a target $\delta$ supplied as input bits, form $y_i=x_i\oplus\delta$ in parallel. The test $y_i<x_i$ is shallow: OR over the possible highest differing bit, requiring equality at every higher bit and $y_{i,k}=0,x_{i,k}=1$ at that bit. Each test has size polynomial in $B$. When no cancelling candidate exists, use the first nonzero heap and subtract one. Bit $k$ of that decrement is $x_{i,k}$ XOR the conjunction saying that all lower bits are zero. The validity bit is false exactly when the board is terminal. In restoration, the target comes from the two-frame delta; in the multi-head readout it comes from reconstruction of the rollout outcomes.

\paragraph{A binary index is part of the circuit.}
Let $s_i$ be the candidate flags after choosing between cancellation and fallback. More than one may be true. Define
\[
f_i=s_i\wedge\bigwedge_{j<i}\neg s_j,
\qquad
I_k=\bigvee_{\substack{0\le i<N\\\text{bit }k\text{ of }i\text{ is }1}}f_i.
\]
The $f_i$ flags select the first candidate; the $I_k$ wires give its binary index. Masking the candidate replacement bits by $f_i$ and taking OR gives the value output. Combining validity with $\bigvee_i s_i$ handles the absence of candidates. This uses constant additional depth and polynomial size, rather than an uncosted external search for the first true flag. The Lean priority encoder proves agreement for every flag pattern, including invalid outputs, no candidates, and the cases $N=0$ and $N=1$. A move index uses $r$ bits with $N\le2^r$; the replay counter also stores $N$ and therefore requires $N<2^r$.

\paragraph{The recurrent update.}
Scan mode updates the accumulator by two-bit XORs. Replay mode tests the current row against the stored summary, retains the first cancelling candidate and a fallback, and increments the row counter. Counter bit $k$ changes exactly when all lower bits are one, so increment also has a shallow circuit. Mode, validity and reset flags select the appropriate register updates. Induction over well-formed token streams proves that every reachable register fits its allotted width and that the output move uses the summary from the same board's first scan. The input protocol supplies the repeated rows; the controller does not secretly store the board.

\paragraph{Realisation with fixed-scale threshold gates.}
An OR gate uses weights $1$ and threshold $1$; a negated literal supplies NOT. Implement AND by De Morgan, as an OR of negated inputs followed by negation. This keeps the threshold bound fixed even when fan-in grows. Constants use no inputs and threshold $0$ or $1$. Thus all these circuits can use the fixed grid $W=D=1$.
More precisely, for $K$ output formulas of depth at most $d$, total formula size $S$, and input support below $n$, the checked compiler constructs one network of at most $2d+2$ layers, at most $(2d+1)S+K$ gates, and layer widths and fan-ins at most $S+n$. Identity buffers align layers; every output is proved equal to its source formula on the same finite input. Size counts duplicated formula nodes, and depth counts NOT gates, so these are conservative bounds.

\begin{figure}[htbp]
\centering
\begin{tikzpicture}[font=\small, box/.style={draw,rounded corners,align=center,text width=3.1cm,minimum height=1.3cm}, >=Stealth]
\node[box] (a) {Boolean formulas\\depth at most $d$};
\node[box,right=0.4cm of a] (b) {Translate gates\\OR/NOT: one layer\\AND: two layers};
\node[box,right=0.4cm of b] (c) {Copy values\\to later layers\\align dependencies};
\node[box,right=0.4cm of c] (d) {Arrange outputs\\at most $2d+2$\\threshold layers};
\draw[->] (a) -- (b); \draw[->] (b) -- (c); \draw[->] (c) -- (d);
\end{tikzpicture}
\caption{The checked Boolean-to-threshold compiler, shown schematically. Negated input literals are permitted. AND uses an OR of negated inputs followed by negation, keeping weights and thresholds in $\{0,1\}$. Copying and output stages give a uniform depth bound.}
\label{fig:threshold-compilation}
\end{figure}

\begin{center}
\begin{tabular}{lcc}
\hline
\textbf{Component} & \textbf{Boolean depth} & \textbf{Threshold layers} \\
\hline
Two-frame binary move selector & $\le20$ & $\le42$ \\
Multi-head binary readout, given outcomes & $\le19$ & $\le40$ \\
One structured policy's binary move output & $\le17$ & $\le36$ \\
One recurrent state update and output & $\le8$ & $\le18$ \\
One shootout controller tick & $\le34$ & $\le70$ \\
\hline
\end{tabular}
\end{center}
These explicit Lean-checked bounds are not optimised. Together with the polynomial size bounds, they certify the components' $\AC^0$ resource requirements. They apply to one evaluation, not to an entire rollout or row scan. The whole-game proofs connect the resulting network outputs to the corresponding game moves and state updates; the time needed to repeat those evaluations remains part of the algorithm.

\paragraph{Where the layer counts come from.}
\label{par:depth-accounting}
For the two-frame selector, the compositional Boolean-depth bound is $4+4+9+3=20$: at most four levels extract the frame difference, four handle candidate tests and replacement bits, nine are allowed by the general selection construction (including fallback and output handling), and three encode the selected heap's binary index. These are bounds on composed subcircuits, not necessarily the exact number of levels on a shortest implementation. The compiler then gives at most $2\cdot20+2=42$ threshold layers. Similarly, Boolean depths $19$, $17$, $8$ and $34$ give the other entries in the table.

The factor of two allows every Boolean level the cost of an AND gate, even though OR and NOT need only one threshold layer. The two additional stages copy inputs into the construction and arrange the output wires; identity gates retain values until they are needed. This uniform construction makes correctness and resource accounting straightforward, but is not a depth optimisation. The table proves that networks within these bounds exist, not that these depths are necessary or that training will discover their weights. A more economical implementation may use fewer layers; establishing sharper bounds would require a separate construction and proof.

\paragraph{Outer-controller storage and time.}
\label{par:controller-costs}
For $B,N\ge1$ and $B\le2^r$, the shootout controller uses $NB+4+r+B$ state bits: a working board, control flags, a head index and the collected outcomes. Its input also contains the original $NB$-bit board, held fixed throughout the decision. This external storage is additional to the state count. The circuit handles head selection, board reload, outcome collection, increment and halt; the depth bound in the table applies to either the early-exit classifier or the all-head schedule.

Let $M$ be the total number of simulated legal moves in the all-head schedule. The implementation uses $M+2B+2$ component evaluations: $M$ move updates, $B$ terminal tests, $B$ outcome-consumption updates, one initial reset and one final move readout. Since $M\le BN(2^B-1)$, this agrees with the rollout-time bound in the main text while making the control overhead explicit. These are circuit evaluations, not measured processor time. The early-exit schedule uses only the visited heads and gives the same classification as the OR of all outcomes. These certificates concern the two specified schedules; the abstract definition of a finite-state controller does not certify arbitrary scheduling logic.

\section{Weak mastery examples}
\label{app:weak}

\begin{example}[Pairing strategy]
Consider positions $(a,a,b,b,c,c,\dots)$ consisting of pairs of equal heaps. The second player can mirror the first player’s move on the matching heap, maintaining the paired structure until termination, hence winning. The checked binary selector finds the first unequal pair and reduces its larger heap to match its partner. It has constant depth and polynomial size, and wins against every legal opponent from the paired family when playing second. For the first-player role every paired initial board is losing, so weak mastery there would be vacuous.
\end{example}

The distinction from strong mastery is substantive. From the single heap $(3)$, a strategy that takes all three stones wins immediately, irrespective of what it would do elsewhere. It therefore weakly masters that initial position as first player. If it has no prescribed move at $(1)$, however, it fails strong mastery: arbitrary legal play can reach $(3)\to(2)\to(1)$ with the first player to move. That history is excluded by its own initial action. The general definition and this separation are checked in Lean.

\section{Proof of Lemma~\ref{lem:nimber-diff-ac0}}
\label{app:nimberdiff}

We provide an explicit $\AC^0$ construction for computing nimber differences between positions differing in at most $t$ heaps.

\begin{proof}[Proof of Lemma~\ref{lem:nimber-diff-ac0}]
Let positions be encoded as $h$ blocks of length $B$ bits (heap sizes in binary), so total input length is $n=hB$.

\textbf{Step 1: identify changed heaps.}
For each heap $i$ and bit position $k\in[B]$, compute $d_{i,k} = x_{i,k}\oplus x'_{i,k}$.
Then $c_i = \bigvee_{k\in[B]} d_{i,k}$ indicates whether heap $i$ changed.
This is depth two after flattening the OR of the two-bit XOR DNFs, ignoring input negations in the usual convention.

\textbf{Step 2: assume at most $t$ changes.}
By promise, $\sum_i c_i \le t$ with $t$ constant.

\textbf{Step 3: compute per-changed-heap XOR values.}
For each heap $i$, define the block XOR value $y_i = x_i \oplus x'_i$ (bitwise XOR across the $B$ bits). Each output bit is XOR of two bits, hence $\AC^0$.

\textbf{Step 4: XOR the at-most-$t$ nonzero blocks.}
Since at most $t$ heaps changed, $\Delta = \bigoplus_{i:c_i=1} y_i$ is XOR of at most $t$ $B$-bit numbers.
The locations of the changed blocks must also be selected. For each $S\subseteq[h]$ with $|S|\le t$, form an indicator that $S$ is exactly the changed set: $\bigwedge_{i\in S}c_i\wedge\bigwedge_{i\notin S}\neg c_i$. Conjoin it with the constant-size truth table for the parity of $\{y_{i,k}:i\in S\}$, and OR over all such $S$. There are $O(h^t)$ choices, each using at most $O(h)$ wires plus a constant-size parity table. This gives polynomial size and constant depth. On the single-change promise, each delta bit simplifies to $\bigvee_i y_{i,k}$.

The overall circuit has constant depth and polynomial size in $n$.
\end{proof}


\section{Independent noise and repetition}
\label{app:noise-samples}

Remark~\ref{rem:noisy-rollout-decay} concerns a symmetric observation channel for an unknown bit $Y$. Each observation agrees with $Y$ with probability $(1+a)/2$, where $0<a\le1/2$, and observations are independent conditional on $Y$. A possibly randomised decision rule assigns a probability of outputting~1 to each observed bit vector. Let $m_*(a)$ be the least sample count for which such a rule has error at most $1/4$ for each possible value of $Y$.

\begin{proposition}[Repetitions needed for a weakly informative bit]
\label{prop:noise-samples}
The minimum $m_*(a)$ exists and satisfies
\[
\frac{3}{32a^2}\le m_*(a)\le\left\lceil\frac{4}{a^2}\right\rceil
\le\frac{17}{4a^2}.
\]
Majority voting, with ties declared~0, achieves the middle upper bound.
\end{proposition}
\begin{proof}[Proof outline]
Under either value of $Y$, the sum of $m$ correctness indicators has mean $m(1+a)/2$ and variance at most $m/4$. The second-moment bound gives majority error at most $1/(ma^2)$, so $m=\lceil4/a^2\rceil$ suffices. This also proves that a least sufficient natural number exists. For necessity, the likelihood ratio between the two product laws has chi-square divergence
\[
\left(1+\frac{4a^2}{1-a^2}\right)^m-1.
\]
If $ma^2<3/32$, use $1-a^2\ge3/4$ and
$(1+u)^m\le(1-mu)^{-1}$ for $u\ge0$, $mu<1$, to bound this divergence strictly below~1. Cauchy--Schwarz then bounds total variation strictly below $1/2$. A rule whose two errors are at most $1/4$ would distinguish the laws with acceptance probabilities differing by at least $1/2$, a contradiction. This applies to randomised rules as well. Finally, $a^2\le1/4$ bounds the ceiling by $4/a^2+1\le17/(4a^2)$.
\end{proof}

For the flip model, $a=(1-2\varepsilon)^L$. Fresh independent blocks of $L$ flips produce exactly the product observation law used above. When $0<\varepsilon<1/2$, the small-bias condition holds for every $L\ge\lceil1/(2\varepsilon)\rceil$; one elementary bound is $(1-2\varepsilon)^L\le(1+2\varepsilon L)^{-1}$. If $\varepsilon=0$ or $L=0$, one observation is exact. If $a=0$, no number of observations permits error below $1/2$ for both ideal bits. Reusing the same corrupted bit leaves the information content unchanged. The finite product laws, real-parameter bounds, minimum sample count and these boundary cases are checked in Lean.

\section{Additional proofs for the rollout analysis}
\label{app:rollout-proofs}

\begin{proof}[Proof of Proposition~\ref{prop:rollout-sufficiency}]
Fix a position $P$ with heap sizes $h_1,\dots,h_N\in\{0,\dots,2^B-1\}$ and let $\sigma=\operatorname{nim}(P)=h_1\oplus\cdots\oplus h_N$.
For each $j\in\{0,\dots,B-1\}$ define the deterministic rollout policy $\pi_j$:
\begin{quote}
``Subtract $2^j$ from the leftmost heap of size at least $2^j$; if none exists, subtract $1$ from the leftmost nonzero heap.''
\end{quote}
Each $\pi_j$ is $\AC^0$-computable because it performs constant-depth comparisons to the threshold $2^j$ and selects the leftmost qualifying heap.

Consider a rollout under $\pi_j$ from $P$.
While some heap is at least $2^j$, every move subtracts exactly $2^j$ from some heap, so the number of such moves is
$L_1=\sum_{i=1}^N \lfloor h_i/2^j\rfloor$.
After this phase all heaps satisfy $h_i<2^j$, and $\pi_j$ falls back to subtracting $1$ until termination, for a further
$L_2=\sum_{i=1}^N (h_i\bmod 2^j)$ moves.
Thus the total number of moves is $L=L_1+L_2$, and (normal play) the first player wins iff $L$ is odd, i.e.\ iff $L\bmod 2=1$.

For each $i$, the parity of $\lfloor h_i/2^j\rfloor$ equals the $j$th bit of $h_i$, denoted $h_i^{(j)}$, so
$L_1\bmod 2=\bigoplus_i h_i^{(j)}=\sigma_j$.
For $j\ge1$, $(h_i\bmod 2^j)\bmod2=h_i^{(0)}$, so $L_2\bmod2=\sigma_0$. For $j=0$, every remainder is zero and $L_2=0$.
Hence
$\mathrm{RO}_{\pi_0}(P)=\sigma_0$ and $\mathrm{RO}_{\pi_j}(P)=\sigma_j\oplus\sigma_0$ for $j\ge 1$.
From these $B$ outcomes the agent recovers $\sigma$.

When $\sigma\ne0$, let $k$ be its highest set bit. Some heap has bit $k$ set; XOR with $\sigma$ changes that bit to zero and leaves every higher bit unchanged, so a decreasing replacement exists. Choose the first such heap and set $h_i'=h_i\oplus\sigma$. The new nim-sum is $\sigma\oplus h_i\oplus h_i'=0$. Comparisons, XOR and first-index selection are $\AC^0$. When $\sigma=0$, output any legal move, or halt if every heap is zero.

This selector gives a winning strategy against every legal opponent response from a nonzero starting position. After the selected move, any opponent move changes one heap by a nonzero XOR difference and therefore makes the nim-sum nonzero again. The selector can restore zero on its next turn. Every legal move strictly decreases the total number of stones, so play terminates; since a nonterminal zero-nim-sum position cannot move directly to the terminal position, the selector makes the last move. The opponent need not follow any rollout policy.

Compute and retain each of the $B$ root outcomes once per decision. Each rollout has at most $\sum_i h_i\le N(2^B-1)$ moves, so their total is at most $B\sum_i h_i\le BN(2^B-1)$. This counts simulated game moves, rather than elementary bit operations or elapsed time. Multiplying by the polynomial gate work per policy update gives a bit-work bound; it is polynomial in $N$ when $B=O(\log N)$, and the displayed step bound is exponential in unrestricted binary heap width.
\end{proof}

\section{Proof of Theorem~\ref{thm:search-gap}}
\label{app:search-gap-proof}

\begin{proof}
\textbf{Part (i).} Fix $B\ge2$, $N\ge1$ and put $q=2^B$. Let $U$ be the uniform distribution on all $N$ independent $B$-bit heaps, before conditioning on a nonzero nim-sum. For any fixed values of $N-1$ heaps and any target nim-sum, exactly one value of the remaining heap gives that target. Thus $\sigma$ is exactly uniform on $\mathbb F_2^B$, and $\Pr_U[\sigma\ne0]=(q-1)/q$.
Let $\ell=\ell_j$ be the nonzero linear functional returned by the fixed structured head, and let $\beta=\ell(\sigma)$.

For each fixed $b\in\{0,1\}$, decode $F_N(P,b)$ as a proposed move $(i,v)$. On legal outputs define $g_b(P)=h_i\oplus v$, and set $g_b(P)=0$ on invalid outputs. Selecting an indexed input block, checking legality and taking its XOR with the output use polynomial-size, constant-depth circuits. Thus for each $a\in\mathbb F_2^B$, the indicator
\[
H_{b,a}(P)=\mathbf1\{g_b(P)=a\}
\]
is in $\AC^0$. On $\sigma\ne0$, a legal move wins exactly when $g_\beta(P)=\sigma(P)$; invalid outputs cannot satisfy that equality.

For $u\in\mathbb F_2^B$, write $\chi_u(P)=(-1)^{\langle u,\sigma(P)\rangle}$. First consider the finite probability calculation under the explicit assumption that some $\eta_N\ge0$ satisfies
\[
\bigl|\mathbb E_U[H_{b,a}(P)\chi_u(P)]\bigr|
\le \eta_N
\]
for every $b\in\{0,1\}$, $a\in\mathbb F_2^B$ and nonzero $u\in\mathbb F_2^B$. This is a two-sided correlation bound under the unconditioned distribution $U$; it does not assume independence between the readout and the nim-sum.
Expanding the exact-value indicator gives
\[
\mathbf1\{\sigma=a\}=q^{-1}\sum_{u\in\mathbb F_2^B}
(-1)^{\langle u,a\rangle}\chi_u(P),
\]
so
\[
\left|\mathbb E_U[H_{b,a}\mathbf1\{\sigma=a\}]
-q^{-1}\mathbb E_U H_{b,a}\right|
\le \frac{q-1}{q}\eta_N.
\]
For each fixed $b$, the events $g_b(P)=a$ are disjoint. Therefore
\begin{align*}
\Pr_U[\text{winning move and }\sigma\ne0]
&=\sum_{a\ne0}\mathbb E_U[H_{\ell(a),a}\mathbf1\{\sigma=a\}]\\
&\le q^{-1}\sum_{b\in\{0,1\}}\sum_{\substack{a\ne0\\\ell(a)=b}}
\mathbb E_U H_{b,a}+\frac{(q-1)^2}{q}\eta_N\\
&\le \frac{2}{q}+\frac{(q-1)^2}{q}\eta_N.
\end{align*}
Dividing by $\Pr_U[\sigma\ne0]=(q-1)/q$ gives the finite bound
\begin{equation}
\Pr_U[\text{legal winning move}\mid\sigma\ne0]
\le \frac{2}{q-1}+(q-1)\eta_N.
\label{eq:finite-correlation-bound}
\end{equation}

It remains to supply the correlation estimate for the circuit class. For $u\ne0$, $\chi_u$ is parity on $N$ times the number of nonzero coordinates of $u$, hence on at least $N$ input bits. Fix all coordinates outside this support. The restricted indicator remains a polynomial-size, fixed-depth $\AC^0$ circuit, so the standard parity-correlation bound applies uniformly to every such restriction; averaging over the fixed coordinates preserves the bound \citep{hastad1986thesis,hastad2014correlation}. Taking the maximum over the finitely many $a,b,u\ne0$ gives a nonnegative negligible $\eta_N$. Since $B$, the circuit depth and the polynomial size exponent are fixed, $(q-1)\eta_N$ is also negligible. This yields the asserted asymptotic bound. The fixed-$B$ assumption is essential to this stated asymptotic conclusion.

For attainability of the leading constant, choose a nonzero $a_b$ from each fibre $\ell(a)=b$, possible because $B\ge2$. After observing $b$, try to cancel $a_b$ using the first heap $h_i$ with $h_i\oplus a_b<h_i$, with any legal fallback if none exists. This is an $\AC^0$ readout and wins whenever $\sigma=a_0$ or $a_1$, an event of conditional probability $2/(q-1)$. Any additional success is at most negligible by the upper bound. For $B=2$ and $\pi_0$, the choices $a_0=2,a_1=1$ give exactly $2/3$: the case $\sigma=3$ is assigned a legal delta-1 move and cannot win.

\textbf{Part (ii).} The $B$ root rollout bits reconstruct $\sigma$. If it is nonzero, a heap containing its highest set bit can be decreased to $h_i\oplus\sigma$, leaving zero nim-sum. An $\AC^0$ first-qualifying-heap selector implements this choice. With zero nim-sum, any legal move is optimal in the game-theoretic sense that no winning move exists. Each root rollout has at most $N(2^B-1)$ moves. Hence $B$ rollouts use $O(BN2^B)$ simulated moves, each with polynomial circuit work; no successor enumeration or unrestricted controller is used.

\textbf{Part (iii).} If the predecessor $Q$ has nim-sum zero and the opponent legally reaches $Q'$, then $\delta=\Delta(Q,Q')=\operatorname{nim}(Q')\ne0$. Select a decreasing heap $h_i\to h_i\oplus\delta$ as above. The two-frame difference and move selector are $\AC^0$, and the response returns nim-sum zero. Repeating this response after the player's own successful moves wins from the established invariant. No assertion about establishing it from an arbitrary start is used.
\end{proof}


\begin{thebibliography}{99}

\bibitem[Ajtai(1993)]{ajtai1993approxcount}
Mikl{\'o}s Ajtai.
\newblock Approximate counting with uniform constant-depth circuits.
\newblock In \emph{Advances in Computational Complexity Theory}, DIMACS Series in Discrete Mathematics and Theoretical Computer Science, vol.~13, pages 1--20, 1993.

\bibitem[Berlekamp et~al.(2001--2004)Berlekamp, Conway, and Guy]{berlekamp2001winningways}
Elwyn~R. Berlekamp, John~H. Conway, and Richard~K. Guy.
\newblock \emph{Winning Ways for Your Mathematical Plays}, Vols.~1--4.
\newblock AK Peters/CRC Press, 2001--2004.

\bibitem[\v{C}{\'i}\v{z}ek et~al.(2026)\v{C}{\'i}\v{z}ek, Balko, and Schmid]{cizek2026sprouts}
Tom{\'a}\v{s} \v{C}{\'i}\v{z}ek, Martin Balko, and Martin Schmid.
\newblock Massively parallel proof-number search for impartial games and beyond.
\newblock \emph{Proceedings of the AAAI Conference on Artificial Intelligence}, 40(43):36838--36846, 2026.
\newblock \url{https://doi.org/10.1609/aaai.v40i43.41010}.

\bibitem[Daniely and Malach(2020)]{daniely2020parities}
Amit Daniely and Eran Malach.
\newblock Learning parities with neural networks.
\newblock In \emph{Advances in Neural Information Processing Systems}, 33, 2020.

\bibitem[Dettmers et~al.(2022)]{dettmers2022llmint8}
Tim Dettmers, Mike Lewis, Younes Belkada, and Luke Zettlemoyer.
\newblock LLM.int8(): 8-bit matrix multiplication for transformers at scale.
\newblock \emph{arXiv preprint arXiv:2208.07339}, 2022.

\bibitem[Friedman(2017a)]{friedman2017challenge}
Harvey Friedman.
\newblock AI Challenge.
\newblock \emph{FOM} mailing list, September 28, 2017.
\newblock \url{https://fomarchive.ugent.be/2017-September/020612.html}.

\bibitem[Friedman(2017b)]{friedman2017deeplearning}
Harvey Friedman.
\newblock NIM/Deep Learning.
\newblock \emph{FOM} mailing list, October 15, 2017.
\newblock Includes correspondence from John Schulman, October 14--15, 2017.
\newblock \url{https://fomarchive.ugent.be/2017-October/020642.html}.

\bibitem[Furst et~al.(1984)Furst, Saxe, and Sipser]{furst1984parity}
Merrick Furst, James~B. Saxe, and Michael Sipser.
\newblock Parity, circuits, and the polynomial-time hierarchy.
\newblock \emph{Mathematical Systems Theory}, 17(1):13--27, 1984.

\bibitem[Hao et~al.(2022)Hao, Angluin, and Frank]{hao2022hardattention}
Yiding Hao, Dana Angluin, and Robert Frank.
\newblock Formal language recognition by hard attention transformers:
perspectives from circuit complexity.
\newblock \emph{Transactions of the Association for Computational Linguistics},
10:800--810, 2022.

\bibitem[H{\aa}stad(1986)]{hastad1986thesis}
Johan H{\aa}stad.
\newblock \emph{Computational Limitations for Small Depth Circuits}.
\newblock PhD thesis, MIT, 1986.

\bibitem[H{\aa}stad(2014)]{hastad2014correlation}
Johan H{\aa}stad.
\newblock On the correlation of parity and small-depth circuits.
\newblock \emph{SIAM Journal on Computing}, 43(5):1699--1708, 2014.
\newblock doi:10.1137/120897432.

\bibitem[Jacob et~al.(2018)]{jacob2018integeronly}
Benoit Jacob, Skirmantas Kligys, Bo Chen, Menglong Zhu, Matthew Tang, Andrew Howard, Hartwig Adam, and Dmitry Kalenichenko.
\newblock Quantization and training of neural networks for efficient integer-arithmetic-only inference.
\newblock In \emph{Proceedings of the IEEE Conference on Computer Vision and Pattern Recognition (CVPR)}, pages 2704--2713, 2018.

\bibitem[Karpov et~al.(2023)Karpov, Markstr{\"o}m, Riis, and Zhou]{karpov2023setalternating}
Alexander Karpov, Klas Markstr{\"o}m, S{\o}ren Riis, and Bei Zhou.
\newblock Set-alternating schemes: A new class of large Condorcet domains.
\newblock \emph{arXiv preprint arXiv:2308.02817}, 2023.

\bibitem[Linial et~al.(1993)Linial, Mansour, and Nisan]{linial1993constant}
Nathan Linial, Yishay Mansour, and Noam Nisan.
\newblock Constant depth circuits, Fourier transform, and learnability.
\newblock \emph{Journal of the ACM}, 40(3):607--620, 1993.

\bibitem[Merrill and Sabharwal(2024)]{merrill2021transformer}
William Merrill and Ashish Sabharwal.
\newblock The expressive power of transformers with chain of thought.
\newblock In \emph{International Conference on Learning Representations}, 2024.
\newblock \url{https://arxiv.org/abs/2310.07923}.

\bibitem[Minsky and Papert(1969)]{minsky1969perceptrons}
Marvin Minsky and Seymour Papert.
\newblock \emph{Perceptrons: An Introduction to Computational Geometry}.
\newblock MIT Press, 1969.

\bibitem[Mnih et~al.(2015)]{mnih2015humanlevel}
Volodymyr Mnih, Koray Kavukcuoglu, David Silver, Andrei~A. Rusu, Joel Veness,
Marc~G. Bellemare, Alex Graves, Martin Riedmiller, Andreas~K. Fidjeland,
Georg Ostrovski, et~al.
\newblock Human-level control through deep reinforcement learning.
\newblock \emph{Nature}, 518(7540):529--533, 2015.

\bibitem[Nesterov(1983)]{nesterov1983method}
Yurii Nesterov.
\newblock A method of solving a convex programming problem with convergence rate $O(1/k^2)$.
\newblock \emph{Soviet Mathematics Doklady}, 27(2):372--376, 1983.

\bibitem[Ngo et~al.(2018)]{ngo2018worstcasejoins}
Hung~Q. Ngo, Ely Porat, Christopher R{\'e}, and Atri Rudra.
\newblock Worst-case optimal join algorithms.
\newblock \emph{Journal of the ACM}, 65(3):16:1--16:40, 2018.

\bibitem[Nowakowski(2025)]{nowakowski2025unsolved}
Richard~J. Nowakowski.
\newblock Unsolved problems in combinatorial games.
\newblock In \emph{Games of No Chance 6}, MSRI Publications, vol.~71, pages 55--98, 2025.
\newblock \url{https://library.slmath.org/books/Book71/files/GONC6-04.pdf}.



\bibitem[Rahaman et~al.(2019)]{rahaman2019spectral}
Nasim Rahaman, Aristide Baratin, Devansh Arpit, et~al.
\newblock On the spectral bias of neural networks.
\newblock In \emph{Proceedings of the 36th International Conference on Machine Learning (ICML)}, pages 5301--5310, 2019.

\bibitem[Razborov(1987)]{razborov1987lower}
Alexander~A. Razborov.
\newblock Lower bounds on the size of bounded depth circuits over a complete basis with logical addition.
\newblock \emph{Mat. Zametki}, 41(4):598--607, 1987.

\bibitem[Riis(2026a)]{riis2026interactive}
S{\o}ren Riis.
\newblock Overcoming a learning bottleneck in impartial games: Interactive edition.
\newblock 2026a.
\newblock \nolinkurl{https://sr123.github.io/impartial-games-learning-lean/}.

\bibitem[Riis(2026b)]{riis2026lean}
S{\o}ren Riis.
\newblock Overcoming a learning bottleneck in impartial games: Lean formalisation and archived interactive edition.
\newblock Version 1.0.0, Zenodo, 2026b.
\newblock \nolinkurl{https://doi.org/10.5281/zenodo.23289503}.
\newblock Lean sources: \nolinkurl{https://github.com/SR123/impartial-games-learning-lean/tree/main/lean}.

\bibitem[Schaefer(1978)]{schaefer1978games}
Thomas~J. Schaefer.
\newblock On the complexity of some two-person perfect-information games.
\newblock \emph{Journal of Computer and System Sciences}, 16(2):185--225, 1978.

\bibitem[Schrittwieser et~al.(2020)]{schrittwieser2020muzero}
Julian Schrittwieser, Ioannis Antonoglou, Thomas Hubert, Karen Simonyan,
Laurent Sifre, Simon Schmitt, Arthur Guez, Edward Lockhart, Demis Hassabis,
Thore Graepel, et~al.
\newblock Mastering atari, go, chess and shogi by planning with a learned model.
\newblock \emph{Nature}, 588(7839):604--609, 2020.

\bibitem[Silver et~al.(2016)]{silver2016alphago}
David Silver, Aja Huang, Chris~J. Maddison, Arthur Guez, Laurent Sifre,
George Van Den~Driessche, Julian Schrittwieser, Ioannis Antonoglou,
Veda Panneershelvam, Marc Lanctot, et~al.
\newblock Mastering the game of go with deep neural networks and tree search.
\newblock \emph{Nature}, 529(7587):484--489, 2016.

\bibitem[Silver et~al.(2017)]{silver2017alphagozero}
David Silver, Julian Schrittwieser, Karen Simonyan, Ioannis Antonoglou,
Aja Huang, Arthur Guez, Thomas Hubert, Lucas Baker, Matthew Lai,
Adrian Bolton, et~al.
\newblock Mastering the game of go without human knowledge.
\newblock \emph{Nature}, 550(7676):354--359, 2017.

\bibitem[Silver et~al.(2018)]{silver2018alphazero}
David Silver, Thomas Hubert, Julian Schrittwieser, Ioannis Antonoglou,
Matthew Lai, Arthur Guez, Marc Lanctot, Laurent Sifre, Dharshan Kumaran,
Thore Graepel, et~al.
\newblock A general reinforcement learning algorithm that masters chess, shogi, and go through self-play.
\newblock \emph{Science}, 362(6419):1140--1144, 2018.

\bibitem[Smolensky(1987)]{smolensky1987algebraic}
Roman Smolensky.
\newblock Algebraic methods in the theory of lower bounds for boolean circuit complexity.
\newblock In \emph{Proceedings of the 19th Annual ACM Symposium on Theory of Computing}, pages 77--82, 1987.

\bibitem[Tal(2015)]{tal2015tight}
Avishay Tal.
\newblock Tight bounds on the Fourier spectrum of $\AC^0$.
\newblock \emph{Electronic Colloquium on Computational Complexity (ECCC)}, TR14-174, 2015.

\bibitem[Thornton(1996a)]{thornton1996bp}
Christopher~James Thornton.
\newblock \emph{Backpropagation can't do parity generalisation}.
\newblock University of Sussex, School of Cognitive and Computing Science, 1996.

\bibitem[Thornton(1996b)]{thornton1996parity}
Chris Thornton.
\newblock Parity: the problem that won't go away.
\newblock In \emph{Conference of the Canadian Society for Computational Studies of Intelligence}, pages 362--374. Springer, 1996.

\bibitem[Xu et~al.(2019)]{xu2019frequency}
Zhi-Qin John Xu, Yaoyu Zhang, and Yanyang Xiao.
\newblock Frequency principle: Fourier analysis sheds light on deep neural networks.
\newblock \emph{arXiv preprint arXiv:1901.06523}, 2019.

\bibitem[Zhou(2024)]{zhou2024phd}
Bei Zhou.
\newblock \emph{Reinforcement Learning for Impartial Games and Complex Combinatorial Optimisation Problems}.
\newblock PhD thesis, Queen Mary University of London, 2024.

\bibitem[Zhou et~al.(2024)Zhou, Markstr{\"o}m, and Riis]{zhou2024cdl}
Bei Zhou, Klas Markstr{\"o}m, and S{\o}ren Riis.
\newblock CDL: A fast and flexible library for the study of permutation sets with structural restrictions.
\newblock \emph{SoftwareX}, 28:101951, 2024.

\bibitem[Zhou and Riis(2023)]{zhou2023parity}
Bei Zhou and S{\o}ren Riis.
\newblock Exploring parity challenges in reinforcement learning through curriculum learning with noisy labels.
\newblock \emph{arXiv preprint arXiv:2312.05379}, 2023.

\bibitem[Zhou and Riis(2026)]{zhou2022impartial}
Bei Zhou and S{\o}ren Riis.
\newblock Impartial games: A challenge for reinforcement learning.
\newblock \emph{Machine Learning}, 115:54, 2026. First preprint 2022.
\newblock \url{https://doi.org/10.1007/s10994-026-06996-1}.

\end{thebibliography}
\end{document}